\documentclass[pmlr,twocolumn,10pt]{jmlr}

\mlhtrack{proceedings}

\newif\iffinal
\finaltrue  % camera-ready

\iffinal
    \ifmlhneedspmlr
      \jmlrvolume{346}
      \jmlryear{2026}
    \fi
    \ifmlhfindings \jmlrproceedings{}{ML4H 2026 - Findings Track}\fi
    \ifmlhdemo     \jmlrproceedings{}{ML4H 2026 - Demo Track}\fi
    \jmlrworkshop{Machine Learning for Health (ML4H) 2026}
\else
    \jmlrproceedings{}{Submitted to ML4H 2026: \mlhtrackname}
    \jmlrworkshop{Machine Learning for Health (ML4H) 2026}
\fi

\usepackage{booktabs}
\usepackage{float}
\usepackage{placeins}   % \FloatBarrier, used in the appendix
\usepackage{array}
\usepackage{adjustbox}
\usepackage{pifont}
\graphicspath{{../v5/}}
\newcommand{\cmark}{\ding{51}}
\newcommand{\xmark}{\ding{55}}
\newcommand{\pmark}{$\circ$}
\newcommand{\rot}[1]{\rotatebox[origin=l]{90}{#1}}
\newcommand{\dash}{\textendash}

\makeatletter
\newenvironment{inlinealgo}{\@twocolumnfalse\begin{algorithm2e}[H]}{\end{algorithm2e}}
\makeatother

\newcommand{\native}{\textsc{n}}
\newcommand{\imputed}{\textsc{i}}

\newcommand{\safedef}[2]{\providecommand{#1}{}\renewcommand{#1}{#2}}
\safedef{\R}{\mathbb{R}}
\safedef{\E}{\mathbb{E}}
\safedef{\Normal}{\mathcal{N}}
\safedef{\Cat}{\mathrm{Cat}}
\safedef{\tr}{\operatorname{tr}}
\safedef{\diag}{\operatorname{diag}}
\safedef{\vecop}{\operatorname{vec}}
\safedef{\blockdiag}{\operatorname{blockdiag}}
\newcommand{\KL}[2]{\mathrm{KL}\!\left(#1\,\middle\|\,#2\right)}
\newcommand{\zerov}[1]{\mathbf{0}_{#1}}
\newcommand{\zerom}[2]{\mathbf{0}_{#1\times#2}}
\newcommand{\ind}{\perp\!\!\!\perp}
\newcommand{\qij}{q_\phi(z_{ij}\mid y_{ij})}
\newcommand{\qZi}{q_\phi(Z_i\mid Y_i)}

\newcommand{\qAi}{q(A_{R,i})}
\newcommand{\pZig}{p(Z_i\mid c_i{=}g,A_{R,i},X_i)}
\DeclareMathOperator{\Ent}{H}
\newcommand{\pijg}{p(z_{ij}\mid c_i{=}g,A_{R,i},x_{S,ij},x_{R,ij})}
\newcommand{\pijc}{p(z_{ij}\mid c_i,A_{R,i},x_{S,ij},x_{R,ij})}

\usepackage{tikz}
\usetikzlibrary{arrows.meta,positioning,fit,backgrounds}
\tikzset{
  latent/.style={circle, draw, minimum size=9mm, inner sep=0pt, font=\large},                 % latent r.v.
  observed/.style={circle, draw, fill=gray!30, minimum size=9mm, inner sep=0pt, font=\large}, % observed r.v.
  covariate/.style={rectangle, draw, minimum size=9mm, inner sep=3pt},           % fixed covariate
  param/.style={inner sep=2pt, font=\large},        % global parameter
  plate/.style={draw, rounded corners, inner sep=10pt},
  arr/.style={-{Stealth[length=5pt]}, thick},
}

\theorembodyfont{\upshape}
\theoremheaderfont{\scshape}
\theorempostheader{:}
\theoremsep{\newline}
\newtheorem*{proposition*}{Proposition}
\newtheorem{assumption}{Assumption}
\title[LCMM-VAE for Scalable Trajectory Phenotyping]{Latent Class Mixed Model Variational Autoencoders for Scalable Phenotyping of Health Trajectories}

\author{%
 \Name{Philom{\`e}ne Letzelter} \Email{philomene.letzelter@inria.fr}\\
 \addr Inria, Universit{\'e} Paris Cit{\'e}, Inserm, HeKA, 75015 Paris, France\\
 \addr Withings, 92130 Issy-les-Moulineaux, France
 \AND
 \Name{Violaine Courrier} \Email{violaine.courrier@withings.com}\\
 \addr Withings, 92130 Issy-les-Moulineaux, France
 \AND
 \Name{Sandrine Katsahian} \Email{sandrine.katsahian@aphp.fr}\\
 \addr Universit{\'e} Paris Cit{\'e}, AP-HP, Inria, Inserm, HeKA, 75015 Paris, France
 \AND
 \Name{Agathe Guilloux} \Email{agathe.guilloux@inria.fr}\\
 \addr Inria, Universit{\'e} Paris Cit{\'e}, Inserm, HeKA, 75015 Paris, France
}

\makeatletter\newcommand{\settabletype}{\def\@captype{table}}\makeatother
\begin{document}

\maketitle

\begin{abstract}
Clustering longitudinal health trajectories requires handling high-dimensional measurements, irregular visit times, and subject-level heterogeneity while retaining clinically interpretable structure. 
Existing methods address this only partially: latent-class mixed models yield an interpretable, covariate-driven partition but do not scale past a handful of markers, while deep time-series clustering scales at the cost of that structure and generally requires a common time grid. We introduce the Latent Class Mixed Model Variational Autoencoder (LCMM-VAE), a deep generative model whose latent prior is a mixture of class-specific linear mixed models, so that each subject's latent trajectory decomposes into its class mean trajectory and a subject-specific deviation. We derive a variational EM algorithm whose class responsibilities, random effects and mixed-model parameters are updated in closed form, and provide identifiability guarantees under stated conditions. Experiments on simulated and real longitudinal cohorts show that LCMM-VAE returns a partition, a reconstruction of incomplete trajectories and interpretable class-level dynamics at once, while remaining competitive with classical and deep baselines.

\end{abstract}

\begin{keywords}
deep clustering, longitudinal data, variational autoencoders, mixed-effects models, latent class models, trajectory clustering
\end{keywords}

\paragraph*{Data and Code Availability}
This work uses the PhysioNet 2012 \citep{Silva2012} and gait \citep{Helwig2016} datasets, freely available on PhysioNet and the UCI Machine Learning Repository. Our code is available at \url{https://github.com/philomeneletzelter/lcmm-vae}.

\paragraph*{Institutional Review Board (IRB)}
This research does not require IRB approval. 

% ======================================================
\section{Introduction}
\label{sec:intro}
% ======================================================

Wearables and electronic health records produce longitudinal cohorts of thousands of individuals followed for years across many markers, within which individuals rarely follow a common course and a population average describes almost no one. Partitioning subjects into classes with distinct trajectories is therefore a natural analysis: it condenses latent structure into a few intelligible types \citep{Jain2010}, and it needs no outcome to supervise against. What makes such a partition clinically usable is that each class can be read as a trajectory: a curve in the units of the markers and covariates that can be evaluated at any time.

Two literatures approach this goal from opposite ends. \emph{Latent-class trajectory models}, including group-based trajectory modelling \citep{Nagin2005}, growth mixture models \citep{Muthen2004} and the latent-class mixed model \citep{ProustLima2017}, assign each subject to a latent class and treat time as a covariate so that irregular and unbalanced visit schedules need no alignment. They decompose each trajectory into \emph{fixed effects}, shared by the population and adjusted by class, and \emph{random effects}, a subject's own deviation from its class trajectory. They do not scale, however: all parameters, whose number grows with classes and markers, are estimated jointly by maximising the marginal likelihood, which limits applications to a handful of markers. \emph{Deep clustering} instead learns a nonlinear representation and a partition jointly \citep{Min2018} and scales naturally, but discards that structure. It consumes a subjects-by-visits-by-markers array on a common grid, which asynchronous multi-year follow-up cannot supply without binning and imputation \citep{MaCRLI2021}. It also admits no covariates, so nothing separates the trajectory shape that defines a subgroup from baseline offsets, device differences, or how far along the process each subject happened to be observed \citep{Madiraju2018, MaDTCR2019, deJong2019}.

We propose the \emph{Latent Class Mixed Model Variational Autoencoder} (LCMM-VAE, \figureref{fig:diag_lcmmvae}), which structures the latent space of a variational autoencoder (VAE) as a mixture of class-specific linear mixed models, extending the single-population prior of the LMM-VAE \citep{Ong2025} to a finite mixture. It combines the interpretability of latent-class mixed models with the scalability of deep clustering. Our contributions are:
\begin{itemize}\itemsep1pt \parskip0pt
  \item \textbf{LCMM-VAE},  a deep generative clustering model returning a partition of the subjects, a reconstruction of their incomplete trajectories, and class trajectories interpretable in terms of the covariates (\sectionref{sec:lcmmvae}). 
  \item \textbf{A variational EM} scheme in which the class responsibilities, the random effects and every mixed-model parameter are updated in closed form, without sampling  (\sectionref{sec:estimation}).
  \item \textbf{Identifiability} of the latent-class trajectories and proportions under stated conditions
  (\sectionref{sec:ident_main}).
  \item \textbf{An empirical evaluation} against statistical, deep clustering and imputation baselines on a simulated cohort and two public longitudinal cohorts (\sectionref{sec:results}).
\end{itemize}

% =====================================================
\section{Related Work}
\label{sec:rw}
% =====================================================
\begin{figure*}[t]
\floatconts
  {fig:diag_lcmmvae}
  {\caption{\textbf{LCMM-VAE overview}. Sparse, irregularly sampled observations (\emph{left}) are encoded into a latent space (\emph{middle}) organised as a Gaussian mixture of class-specific linear mixed models, then decoded (\emph{right}) into complete trajectories (dotted: beyond a marker's visits; dashed: absent markers).}}
  {\includegraphics[width=1\linewidth]
  {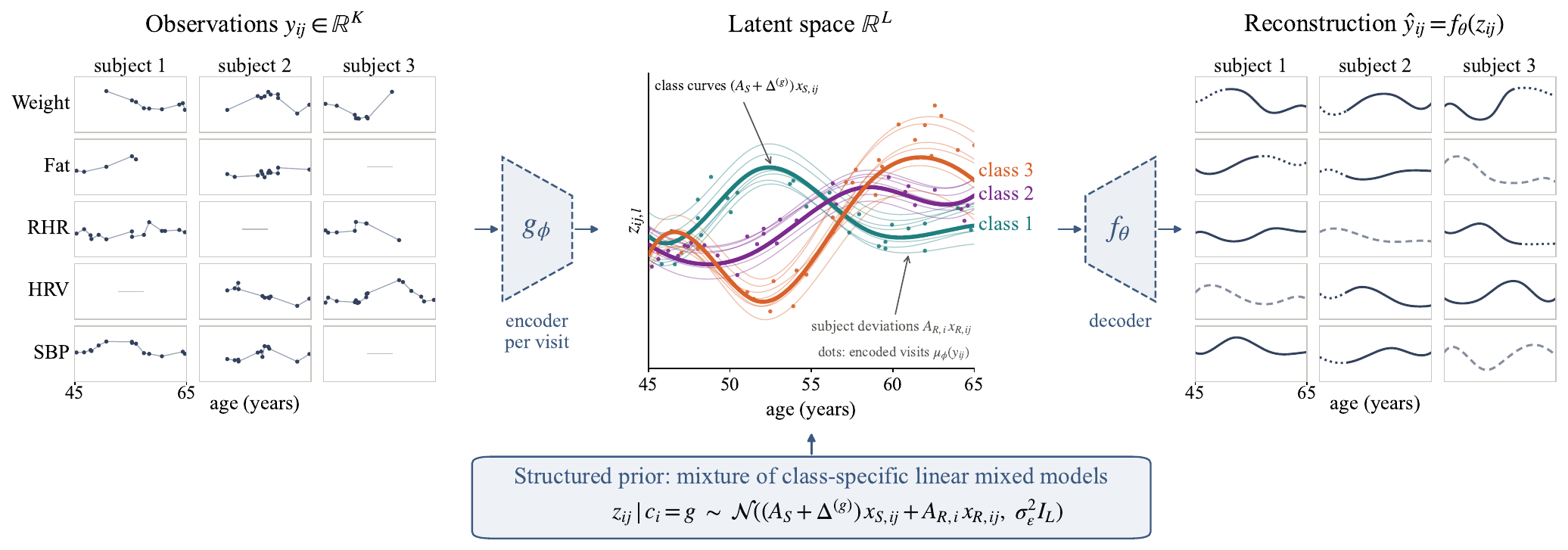}}
\end{figure*}

\paragraph{Trajectory clustering in statistics.}
\label{sec:bg_stats}
Mixed-effects models split a repeated measurement into a population trend and individual deviations \citep{LairdWare1982}. The latent-class models of \sectionref{sec:intro} let this trend differ across classes; the latent-class mixed model (LCMM) \citep{ProustLima2006, ProustLima2017} also links a continuous-time latent process to each marker through a parametric measurement model, but scales to only a few markers. Neighbouring families gain flexibility by leaving the mixed-model form. Functional data clustering fits each subject's curves first with splines \citep{Abraham2003} or functional principal components \citep{Ramsay2005}, and then clusters the components. It scales but is two-stage: the curves are fitted independently of the partition, which needs dense visits, and it models no covariates. Mixtures of Gaussian processes handle irregular sampling at cubic cost per subject \citep{Shi2008}. Disease progression models and their mixtures fit nonlinear mixed effects on Riemannian manifolds \citep{Schiratti2017, Kaisaridi2026}, but assume monotone progression. None learns a shared representation of many markers.

\paragraph{Deep clustering of time series.}
\label{sec:bg_deep}
Deep clustering learns the representation and the partition together. DEC \citep{Xie2016} and IDEC \citep{Guo2017} fine-tune an autoencoder towards learned centroids. VaDE \citep{Jiang2017} and GMVAE \citep{Dilokthanakul2016} instead place a Gaussian-mixture prior on the latent space, drawing a class per data point. Time-series variants add a $k$-means objective and an auxiliary discrimination task to a sequence autoencoder \citep[DTCR;][]{MaDTCR2019}, run VaDE through a long short-term memory network with implicit imputation \citep[VaDER;][]{deJong2019}, optimise imputation and clustering jointly \citep[CRLI;][]{MaCRLI2021}, or assign each visit to a node of a self-organising map \citep{Fortuin2019, Manduchi2021}. None admits covariates, so a class is a free point in a representation space rather than a function of interpretable quantities, and all require a common grid.

\paragraph{Structured-prior VAEs.}
\label{sec:bg_vae}
A third line keeps the amortised VAE \citep{Kingma2014} and replaces the standard normal prior with one that encodes structure, as we do. Temporal structure enters through latent state-space dynamics \citep{Chung2015, Krishnan2017}, latent differential equations \citep{Rubanova2019}, and Gaussian-process priors over time and covariates \citep{Casale2018, Fortuin2020, Ramchandran2021}. The latter scale cubically with the number of observations, unless approximated by inducing points \citep{Jazbec2021} or a spectral basis \citep{Balik2025}. The LMM-VAE \citep{Ong2025}, which we build on, makes the prior a linear mixed model: a regression on the covariates with population and subject effects. Among these temporal priors, only the switching linear dynamical system prior of \citet{Johnson2016} adds discrete states, but these segment a sequence rather than partition subjects.

\paragraph{Relation to prior work.} \appendixref{app:compare} tabulates, for each of these methods, five properties that the setting of \sectionref{sec:intro} calls for: a partition of the subjects, irregular visit times, many markers, covariates with individual deviations, and trajectories in continuous time. LCMM-VAE is the only one that meets all five, and altering one of its components recovers each line of work in turn: a single class ($G{=}1$) gives an LMM-VAE; dropping the covariate design and the random effects gives a VaDE whose class is drawn per subject; and replacing the decoder by per-marker parametric links on a scalar latent process gives the classical LCMM.

\section{The LCMM-VAE}
\label{sec:lcmmvae}

\paragraph{Problem formulation.}
We observe $N$ subjects. Subject $i$ is seen at $n_i$ visits, at times $t_{ij}$, $j=1,\dots,n_i$, whose number and timing differ across subjects. At each visit, $K$ markers $y_{ij}\in\R^{K}$ are recorded, some of which may be missing, as flagged by a mask $o_{ij}\in{\{0,1\}}^{K}$. Each visit also carries covariates: a fixed-effect design $x_{S,ij}\in\R^{S}$, and a random-effect design $x_{R,ij}\in\R^{R}$. Their entries are a fixed basis of time evaluated at $t_{ij}$, possibly with time-invariant attributes. The task is to partition the subjects into $G$ classes, to reconstruct each subject's trajectory at any time, and to describe each class by a trajectory. Classes and trajectories are defined on a latent state $z_{ij}\in\R^{L}$, $L\le K$, a low-dimensional summary of the subject at $t_{ij}$. We stack a subject's visits into $Y_i$ and $X_i$. \appendixref{app:notation} lists every symbol.

\subsection{Generative Model}
\label{sec:lcmm_gen}

The model follows a three-level hierarchy, shown as a plate diagram in \figureref{fig:plate}. First, each subject draws a class $c_i\in\{1,\dots,G\}$ and a matrix of random effects $A_{R,i}\in\R^{L\times R}$, its own coefficients on $x_{R,ij}$,
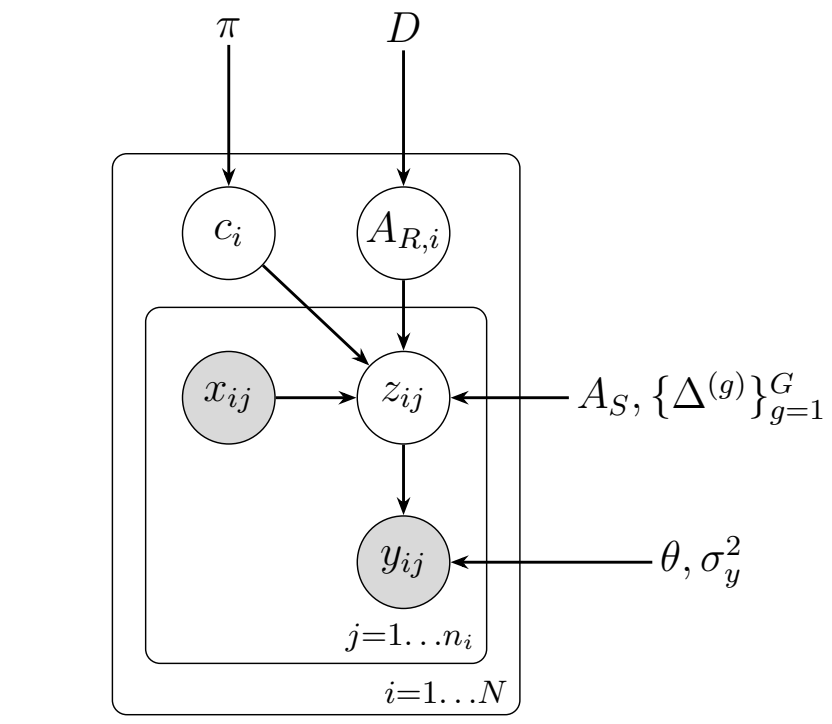
\begin{figure}[t]
\floatconts
  {fig:plate}
  {\caption{\textbf{Plate diagram}. Shaded and blank circles are observed and latent variables, respectively; unframed symbols
  are global parameters, and $x_{ij}=(x_{S,ij},x_{R,ij})$.}}
  {\scalebox{0.68}{\begin{tikzpicture}
    \node[latent]                (ci)  at (0.0, 1.6) {$c_i$};
    \node[latent]                (ARi) at (1.7, 1.6) {$A_{R,i}$};
    \node[observed]              (xij) at (0.0, 0.0) {$x_{ij}$};
    \node[latent]                (zij) at (1.7, 0.0) {$z_{ij}$};
    \node[observed]              (yij) at (1.7,-1.6) {$y_{ij}$};
    \node[param]                 (pi)    at (0.0, 3.6) {$\pi$};
    \node[param]                 (D)     at (1.7, 3.6) {$D$};
    \node[param]                 (AS)    at (4.6, 0.0)
         {$A_S,\{\Delta^{(g)}\}_{g=1}^{G}$};
    \node[param]                 (theta) at (4.6,-1.6) {$\theta,\sigma_y^{2}$};
    \begin{scope}[on background layer]
      \coordinate (padv) at ([yshift=-3pt]yij.south);
      \node[plate, fit=(xij)(zij)(yij)(padv), inner ysep=12pt, inner xsep=10pt,
            label={[anchor=south east, font=\small]south east:$j{=}1{\ldots}n_i$}] (visitplate) {};
      \coordinate (pad) at ([yshift=-5pt]visitplate.south);
      \node[plate, fit=(ci)(ARi)(visitplate)(pad), inner sep=9pt,
            label={[anchor=south east, font=\small]south east:$i{=}1{\ldots}N$}] (subjectplate) {};
    \end{scope}
    \draw[arr] (pi)    -- (ci);
    \draw[arr] (D)     -- (ARi);
    \draw[arr] (AS)    -- (zij);
    \draw[arr] (ci)    -- (zij);
    \draw[arr] (ARi)   -- (zij);
    \draw[arr] (xij)   -- (zij);
    \draw[arr] (zij)   -- (yij);
    \draw[arr] (theta) -- (yij);
  \end{tikzpicture}}}
\end{figure}
\begin{align}
  p_\pi(c_i) &= \Cat(\pi),\label{eq:lcmm_c}\\
  p_{D_l}(a_{R,i,l}) &= \Normal\bigl(\zerov{R},D_l\bigr),\label{eq:lcmm_re}
\end{align}
independently across subjects and latent coordinates, with $a_{R,i,l}^{\top}$ the $l$-th row of $A_{R,i}$. These random effects set the subject's deviation from its class trajectory, with covariance $D_l\in\R^{R\times R}$ shared across subjects; we write $D:=\blockdiag(D_1,\dots,D_L)$. Then at each visit, the latent state is drawn about a class- and subject-specific mean,
\begin{multline}\label{eq:lcmm_z}
  p_{A_S,\Delta^{(g)},\sigma_\varepsilon^{2}}
    \bigl(z_{ij}\mid c_i{=}g,A_{R,i},x_{S,ij},x_{R,ij}\bigr)\\
  = \Normal\bigl(\bar z^{(g)}_{ij},\sigma_\varepsilon^{2}I_L\bigr),
\end{multline}
with latent residual variance $\sigma_\varepsilon^{2}$, fixed to $1$ to set the scale of the latent space, as in \citet{Ong2025}. This mean splits into population, class and subject parts:
\begin{equation}\label{eq:partition}
  \bar z^{(g)}_{ij}
=\underbrace{A_S\,x_{S,ij}}_{\text{population}}
+\underbrace{\Delta^{(g)}x_{S,ij}}_{\text{class $g$}}
+\underbrace{A_{R,i}\,x_{R,ij}}_{\text{subject $i$}},
\end{equation}
where $A_S\in\R^{L\times S}$ is the shared coefficient matrix and $\Delta^{(g)}\in\R^{L\times S}$ the departure of class $g$ from it. Finally, each visit is decoded through a Gaussian likelihood with diagonal covariance:
\begin{equation}\label{eq:lcmm_y}
 p_{\theta,\sigma_y^{2}}\bigl(y_{ij}\mid z_{ij}\bigr)
  = \Normal\bigl(f_\theta(z_{ij}),\diag(\sigma_y^{2})\bigr),
\end{equation}
with $f_\theta:\R^{L}\to\R^{K}$ the decoder and $\sigma_y^{2}\in\R^{K}_{>0}$ learnable per-marker noise variances. The generative parameters are $\Theta=(\pi,\{D_l\},A_S,\{\Delta^{(g)}\},\theta,\sigma_y^{2})$.

\paragraph{Joint density.}
Collecting \eqref{eq:lcmm_c}--\eqref{eq:lcmm_y}, with the latent states stacked into $Z_i$ and $X_i=(X_{S,i},X_{R,i})$, the joint density of one subject is
\begin{multline}\label{eq:joint}
  p\bigl(Y_i,Z_i,A_{R,i},c_i\mid X_i\bigr)
  = p_\pi(c_i)\,p_D(A_{R,i})\\
\times\prod_{j=1}^{n_i}\pijc\;p_{\theta,\sigma_y^2}(y_{ij}\mid z_{ij}),
\end{multline}
with $p_D(A_{R,i})=\prod_l p_{D_l}(a_{R,i,l})$. The decoder is class-agnostic, hence all heterogeneity and all temporal structure live in the prior mean \eqref{eq:partition}: classes differ in \emph{where} they place a trajectory in latent space, never in \emph{how} latent states map to markers, as in the classical LCMM. Finally, $A_S$ and $\Delta^{(g)}$ enter only through $B^{(g)}:=A_S+\Delta^{(g)}$, so we centre the class departures,
\begin{equation}\label{eq:centering}
\textstyle\sum_{g}\pi_g\,\Delta^{(g)}=\zerom{L}{S},
\end{equation}
making $A_S$ the prevalence-weighted mean of the class coefficients, as identifiability requires (\sectionref{sec:ident_main}).

\subsection{Inference Model}
\label{sec:lcmm_inference}
We approximate the intractable posterior $p_\Theta(Z_i,A_{R,i},c_i\mid Y_i,X_i)$ by amortised variational inference \citep{Kingma2014} under the mean-field factorisation
\begin{multline}\label{eq:qfam}
  q_\psi\bigl(Z_i,A_{R,i},c_i\mid Y_i\bigr)
  = \Bigl(\prod_{j=1}^{n_i}q_\phi\bigl(z_{ij}\mid y_{ij}\bigr)\Bigr)\\
  \times\;q_{\gamma_i}(c_i)\; \prod_{l=1}^{L}q_{m_{i,l},V_{i,l}}\bigl(a_{R,i,l}\bigr),
\end{multline}
whose three factors are
\begin{align*}
  q_\phi\bigl(z_{ij}\mid y_{ij}\bigr)
    &= \Normal\bigl(\mu_\phi(y_{ij}),
       \diag(\sigma^{2}_\phi(y_{ij}))\bigr),\\
  q_{m_{i,l},V_{i,l}}\bigl(a_{R,i,l}\bigr)
    &= \Normal\bigl(m_{i,l},V_{i,l}\bigr),
       \\
  q_{\gamma_i}(c_i) &= \Cat(\gamma_i).
\end{align*}
Here $\gamma_i$ holds subject $i$'s class probabilities, the \emph{responsibilities}, $m_{i,l}\in\R^{R}$ is the posterior mean approximating subject $i$'s deviation from its class trajectory on latent coordinate $l$, and $V_{i,l}\in\R^{R\times R}$ the posterior covariance. The inference parameters are thus $\psi=(\phi,\{\gamma_i,m_{i,l},V_{i,l}\}_{i,l})$. Only the encoder weights $\phi$ are amortised, through an inference network $g_\phi:\R^{K}\to\R^{2L}$. The local parameters are set in closed form (\sectionref{sec:estimation}).

\subsection{Evidence Lower Bound}
\label{sec:lcmm_elbo}
We maximise each subject's evidence lower bound (ELBO)
\begin{equation}\label{eq:elbo}
\begin{split}
  \mathcal{L}_i=\ &\underbrace{\sum_{j=1}^{n_i}\E_{\qij}
        \bigl[\log p_{\theta,\sigma_y^{2}}(y_{ij}\mid z_{ij})\bigr]}
        _{\text{(i) reconstruction}}
     -\underbrace{\sum_{g=1}^{G}\gamma_{ig}\kappa_{ig}}
        _{\text{(ii) latent reg.}}\\[2pt]
    &-\underbrace{\KL{\gamma_i}{\pi}}_{\text{(iii) class reg.}}
     -\underbrace{\KL{q(A_{R,i})}{p_{D}(A_{R,i})}}
        _{\text{(iv) random-effect reg.}},
\end{split}
\end{equation}
where
\begin{multline}\label{eq:Kig}
  \kappa_{ig} := \sum_{j=1}^{n_i}
    \E_{q(A_{R,i})}\Bigl[\mathrm{KL}\bigl(\qij\;\big\|\\
    \pijg\bigr)\Bigr]\;\in\R,
\end{multline}
is the cost of placing subject $i$ in class $g$, pooled over that subject's visits. Term~(i) is the reconstruction likelihood, evaluated only on observed entries through the mask $o_{ij}$ \citep{Nazabal2020}; (ii) replaces the VAE's latent divergence by a $\gamma$-weighted sum over the $G$ components; (iii) regularises the assignments towards the mixing proportions; and (iv) shrinks each subject's deviation at a rate set by $D$. \appendixref{app:elbo} derives \eqref{eq:elbo} and gives (ii)--(iv) in closed form. 

\subsection{Identifiability}
\label{sec:ident_main}
Identifiability matters here because the parameters at stake are the ones to be interpreted: the departures $\Delta^{(g)}$ define the classes and the $\pi_g$ their prevalences. Under a full-rank condition on the designs, distinctness of the $\pi_g$ and of the $\Delta^{(g)}$, the existence of one sufficiently well-observed subject, the centring constraint \eqref{eq:centering}, and a continuous, injective decoder, the parameters $\bigl(A_S,\{\Delta^{(g)}\},\{D_l\},\pi\bigr)$ are uniquely determined by the law of the observations, up to a permutation of the class labels, with $f_\theta$ and $\sigma_y^{2}$ held fixed. With a learned decoder, \citet{Kivva2022} show that, if the decoder is piecewise-affine, continuous and injective and the noise has a known distribution, a Gaussian-mixture latent and the decoder are jointly identified up to one invertible affine map. Applied to our model, this would identify the proportions, the partition and the decoded class trajectories up to a class permutation, and $A_S$, the $\Delta^{(g)}$ and the $D_l$ only up to an affine map of the latent space. Both results need $K\ge L$, which holds on every dataset here. \appendixref{app:identifiability} provides the proofs for the fixed decoder (Proposition~\ref{prop:obs}), the argument for the learned one (Remark~\ref{rem:learned}) and what each condition costs (Remark~\ref{rem:cost}). To our knowledge, no deep clustering method for longitudinal data establishes identifiability of its classes.

\section{Training and Inference}
\label{sec:estimation}
Training maximises the bound \eqref{eq:elbo} in three stages, described below. \appendixref{app:mstep} derives every update and gives the full algorithm.

\subsection{Pretraining and Initialisation}
\label{sec:init}
We pretrain a single-class version of the model without random effects, and keep its encoder, decoder and $\sigma_y^2$. The mixture is then initialised on the encoder means by em-EM \citep{Biernacki2003}: several short EM runs, each started from the per-subject regressions of $G$ random subjects, of which the run of highest likelihood is continued.

\subsection{Variational EM}
\label{sec:em}
Training is a variational EM \citep{NealHinton1998} whose updates are in closed form, except for the networks and $\sigma_y^2$, which take gradient steps within the M-step.
  
\paragraph{E-step.}
The E-step updates each subject's random effects and class responsibilities. The posterior of the random effects, $(m_{i,l},V_{i,l})$, is a Bayesian regression on $x_R$ of the subject's encoder means around its class curves: as in a classical mixed model, a subject with few visits is shrunk towards its class. The responsibilities are
\begin{equation}\label{eq:gammastar}
  \gamma_{ig}
  = \frac{\pi_g\exp(-\kappa_{ig})}
         {\sum_{g'}\pi_{g'}\exp(-\kappa_{ig'})},
\end{equation}
the prevalence of class $g$ tilted by the cost $\kappa_{ig}$. Since $\kappa_{ig}$ is a divergence between Gaussians, \eqref{eq:gammastar} is exact given $m_{i,l}$ and involves no sampling. Mixture-prior VAEs instead set this divergence to zero \citep{Jiang2017}, which is unattainable in general, or estimate its optimum by sampling $Z_i$ \citep{Falck2021, Courrier2026}; \appendixref{app:qc} shows that \eqref{eq:gammastar} computes that optimum exactly.

\paragraph{M-step.}
The M-step then updates the prior parameters in closed form: $\pi$ and $D_l$ are the mean responsibilities and the mean second moment of the random effects, and each $B^{(g)}$ is a $\gamma$-weighted least-squares fit of the encoder means on $x_S$, split into $A_S$ and $\Delta^{(g)}$ by \eqref{eq:centering}. With these fixed, one gradient epoch, minibatched over visits, updates $(\phi,\theta,\sigma_y^2)$. Training stops once fewer than a set fraction of subjects change class between successive E-steps \citep{Xie2016}, or after a maximum number of epochs.

\subsection{Inference}
\label{sec:inference}
At inference, an E-step on a subject's observed visits gives its responsibilities $\gamma_i$ and random effects $M_i$ (rows $m_{i,l}^{\top}$). The subject is then decoded at any design row,
\begin{equation}\label{eq:hybrid}
    \hat y_{ij}=f_\theta\Bigl(\textstyle\sum_g\gamma_{ig}B^{(g)}x_{S,ij}+M_ix_{R,ij}\Bigr),
\end{equation}
the latent class means, weighted by the responsibilities, plus the subject's inferred deviation. 

% =========================================================================
\section{Experimental Setup}
\label{sec:experiments}
\label{sec:setup}
% =========================================================================

We evaluate clustering and reconstruction on three cohorts: a simulation, a braced-gait study \citep{Helwig2016} and the PhysioNet/CinC 2012 Challenge \citep{Silva2012}. \appendixref{app:exp_det} details the datasets, the configurations of every method and the compute.

\subsection{Datasets}
\label{sec:datasets}
\paragraph{Simulation.}\label{sec:exp_sim}
To evaluate the model with many markers, sparse visits at subject-specific times and subject-level deviations, we construct a simulated dataset. Three classes follow latent trajectories differing in the timing of a rise, the timing of a transient peak and the rate of a decline. Subjects deviate from their class through a random intercept and slope ($R{=}2$) and smooth subject-level noise. A frozen random $\tanh$ network maps this $3$-dimensional latent state to $10$ informative markers, and $40$ class-independent nuisance markers complete the panel ($K=50$). The generator's decoder, its noise and the nuisance markers lie outside our model. Each subject keeps about $20$ of $100$ visits, dropped completely at random (MCAR) or depending on the last observed value of an informative marker (MAR). We generate $N=500$ subjects and hold out $15\%$.

\paragraph{Gait.}
\label{sec:exp_gait}
The gait study \citep{Helwig2016} records $10$ subjects under $3$ conditions (unbraced, knee brace, ankle brace), $10$ gait cycles each. Each walk (one cycle) records $6$ joint angles at $101$ normalised time points, giving $300$ walks labelled by condition. Each seed holds out two subjects. We additionally impose missingness: \texttt{mcar} hides $20$, $40$ or $60\%$ of cells, leaving hidden cells observed neighbours in time; \texttt{channel} removes whole markers from a walk at rate $p=0.17$, $0.33$ or $0.50$, leaving none.

\paragraph{PhysioNet.}\label{sec:exp_physionet_data}
The PhysioNet/CinC 2012 Challenge \citep{Silva2012} monitors ICU stays over their first $48$\,h, with $37$ time-series variables. We use the extraction of \citet{Fortuin2020}, in which each stay is a $48\times35$ hourly panel with $82\%$ of cells missing.

\subsection{Benchmark}
\label{sec:baselines}
For clustering, we run the related methods that have public code and return an unsupervised partition, directly or by $k$-means on a learned representation. We also include two-stage functional clustering: per-subject B-spline coefficients clustered by $k$-means \citep{Abraham2003} or a diagonal Gaussian mixture (GMM-diag), and FPCA scores clustered by $k$-means \citep{Ramsay2005}. For reconstruction, we add mean fill, last observation carried forward (LOCF), linear interpolation, BRITS \citep{Cao2018} and SAITS \citep{Du2023}. Two ablations isolate the random effects ($R{=}0$) and the mixture ($G{=}1$). LCMM-VAE ($G{=}1$) is our model with a single class, distinct from the LMM-VAE \citep{Ong2025}; \appendixref{app:lmmvae} relates the two.

\begin{table*}[t]
\floatconts
  {tab:main}
  {\caption{\textbf{Clustering}, ARI on held-out subjects, $\text{mean}_{\pm\text{s.d.}}$ over seeds. Bold: best mean per column. Gait: mean over the three rates of each mechanism. \emph{s}: median seconds per fit on the simulation. $^{\dag}$Not run: $R{=}0$ on gait. $^{\ddag}$Infeasible at $K=50$; on gait, no fit within $6$\,h.}}
  {\footnotesize\setlength{\tabcolsep}{5pt}
  \begin{adjustbox}{max width=\linewidth}
  \begin{tabular}{@{}l cc ccc r@{}}
\toprule
 & \multicolumn{2}{c}{Simulation (ARI $\uparrow$)} & \multicolumn{3}{c}{Gait (ARI $\uparrow$)} & \\
\cmidrule(lr){2-3}\cmidrule(lr){4-6}
Method & MCAR & MAR & complete & \texttt{mcar} & \texttt{channel} & s \\
  \midrule
  \multicolumn{7}{@{}l}{\itshape Two-stage functional: partition and class mean curves} \\
  $k$-means (B-spline) \citep{Abraham2003} & $0.68_{\pm 0.16}$ & $0.58_{\pm 0.18}$ & $0.90_{\pm 0.14}$ & $0.90_{\pm 0.11}$ & $0.23_{\pm 0.07}$ & 0.8             \\
  GMM-diag (B-spline) & $0.71_{\pm 0.14}$ & $0.44_{\pm 0.14}$ & $0.97_{\pm 0.10}$ & $\mathbf{0.95}_{\pm 0.08}$ & $0.11_{\pm 0.04}$ & 1.5             \\
  FPCA $+$ $k$-means \citep{Ramsay2005} & $0.28_{\pm 0.07}$ & $0.28_{\pm 0.06}$ & $0.94_{\pm 0.12}$ & $0.91_{\pm 0.12}$ & $0.45_{\pm 0.07}$ & 1.2             \\
  \midrule
  \multicolumn{7}{@{}l}{\itshape Partition only} \\
  IDEC \citep{Guo2017} & $0.13_{\pm 0.13}$ & $0.01_{\pm 0.04}$ & $0.83_{\pm 0.18}$ & $0.73_{\pm 0.12}$ & $\mathbf{0.52}_{\pm 0.10}$ & 34.5             \\
  VaDE \citep{Jiang2017} & $0.00_{\pm 0.00}$ & $0.01_{\pm 0.05}$ & $0.68_{\pm 0.26}$ & $0.58_{\pm 0.15}$ & $0.33_{\pm 0.13}$ & 138             \\
  \midrule
  \multicolumn{7}{@{}l}{\itshape Partition and reconstruction} \\
  VaDER \citep{deJong2019} & $0.03_{\pm 0.06}$ & $0.03_{\pm 0.07}$ & $0.23_{\pm 0.22}$ & $0.16_{\pm 0.15}$ & $0.10_{\pm 0.13}$ & 88.1             \\
  CRLI \citep{MaCRLI2021} & $0.02_{\pm 0.03}$ & $0.02_{\pm 0.05}$ & $0.73_{\pm 0.30}$ & $0.75_{\pm 0.27}$ & $0.46_{\pm 0.17}$ & 2856             \\
  \midrule
  \multicolumn{7}{@{}l}{\itshape Reconstruction, partition post hoc} \\
  GP-VAE \citep{Fortuin2020} $+$ $k$-means & $0.24_{\pm 0.10}$ & $0.08_{\pm 0.08}$ & $0.81_{\pm 0.20}$ & $0.52_{\pm 0.17}$ & $0.40_{\pm 0.12}$ & 56.8             \\
  LCMM-VAE ($G{=}1$, ablation) & $0.88_{\pm 0.08}$ & $0.69_{\pm 0.19}$ & $\mathbf{1.00}_{\pm 0.00}$ & $0.65_{\pm 0.08}$ & $0.42_{\pm 0.08}$ & 11.6             \\
  \midrule
  \multicolumn{7}{@{}l}{\itshape Partition, reconstruction and class trajectories} \\
  LCMM$^{\ddag}$ \citep{ProustLima2017} & \dash & \dash & \dash & \dash & \dash & \dash  \\
  LCMM-VAE (ours, $R{=}0$) & $0.29_{\pm 0.10}$ & $0.27_{\pm 0.08}$ & $0.94_{\pm 0.10}$ & $0.84_{\pm 0.13}$ & $0.51_{\pm 0.09}$ & 27.1  \\
  \textbf{LCMM-VAE (ours, $R{=}2$)} & $\mathbf{0.90}_{\pm 0.08}$ & $\mathbf{0.76}_{\pm 0.20}$ & \dash\rlap{$^{\dag}$} & \dash\rlap{$^{\dag}$} & \dash\rlap{$^{\dag}$} & 12.7  \\
\bottomrule
\end{tabular}
\end{adjustbox}}
\end{table*}

\subsection{Protocol and Metrics}
\label{sec:protocol}
\paragraph{Protocol.} Baselines run their authors' code or a standard library at their published or default configuration. Every method is given the same number of classes $G$: the reference one where it exists, and $G=4$ on PhysioNet, the number of ICU types. In practice, $G$ would be selected by a likelihood criterion, such as BIC or ICL. We run $30$ seeds ($10$ for methods slower than $500$\,s per fit). A seed fixes the data split, the mask and the initialisation, shared by every method.

\paragraph{Metrics.} Partitions are scored by the adjusted Rand index (ARI) of held-out subjects against the reference partition. Reconstructions are scored by the mean squared error (MSE) at hidden cells, on standardised markers. Each method is compared with LCMM-VAE by a two-sided paired Wilcoxon signed-rank test over the shared seeds \citep{Demsar2006}.

\begin{table*}[t]
\floatconts
  {tab:recon}
  {\caption{\textbf{Reconstruction}, MSE at hidden cells of held-out subjects, $\text{mean}_{\pm\text{s.d.}}$ over seeds. Bold: best mean per column. Simulation: the $10$ informative markers. Gait: mean over the three rates of each mechanism. \emph{s}: median seconds per fit on the simulation; n.r.: run on GPU, not reported. $^{\dag}$Not run: $R{=}0$ on gait.}}
  {\footnotesize\setlength{\tabcolsep}{5pt}
    \begin{tabular*}{\linewidth}{@{}l@{\extracolsep{\fill}}cc cc r@{}}
  \toprule
   & \multicolumn{2}{c}{Simulation (MSE $\downarrow$)} & \multicolumn{2}{c}{Gait (MSE $\downarrow$)} & \\
  \cmidrule(lr){2-3}\cmidrule(lr){4-5}
  Method & MCAR & MAR & \texttt{mcar} & \texttt{channel} & s \\
  \midrule
    \multicolumn{6}{@{}l}{\itshape No learned model} \\
    Mean fill & $1.014_{\pm 0.035}$ & $1.211_{\pm 0.101}$ & $1.025_{\pm 0.066}$ & $1.027_{\pm 0.071}$ & 0.6 \\
    LOCF & $0.740_{\pm 0.030}$ & $0.869_{\pm 0.070}$ & $0.049_{\pm 0.007}$ & $1.027_{\pm 0.071}$ & 0.5 \\
    Linear interp. & $\mathbf{0.557}_{\pm 0.023}$ & $\mathbf{0.617}_{\pm 0.044}$ & $\mathbf{0.004}_{\pm 0.001}$ & $1.027_{\pm 0.071}$ & 0.4 \\
    \midrule
    \multicolumn{6}{@{}l}{\itshape Reconstruction, no partition or post hoc} \\
    SAITS \citep{Du2023} & $0.778_{\pm 0.034}$ & $1.112_{\pm 0.097}$ & $0.017_{\pm 0.004}$ & $0.640_{\pm 0.071}$ & n.r. \\
    BRITS \citep{Cao2018} & $0.622_{\pm 0.045}$ & $0.794_{\pm 0.100}$ & $0.007_{\pm 0.001}$ & $0.602_{\pm 0.106}$ & 3490 \\
    GP-VAE \citep{Fortuin2020} & $0.754_{\pm 0.054}$ & $0.819_{\pm 0.066}$ & $0.210_{\pm 0.035}$ & $0.470_{\pm 0.039}$ & 56.8 \\
    LCMM-VAE ($G{=}1$, ablation) & $0.687_{\pm 0.031}$ & $0.801_{\pm 0.069}$ & $0.371_{\pm 0.027}$ & $\mathbf{0.338}_{\pm 0.034}$ & 11.6 \\
    \midrule
    \multicolumn{6}{@{}l}{\itshape Partition, reconstruction and class trajectories} \\
   LCMM-VAE (ours, $R{=}0$) & $0.816_{\pm 0.038}$ & $0.949_{\pm 0.088}$ & $0.277_{\pm 0.022}$ & $0.353_{\pm 0.039}$ & 27.1 \\
  \textbf{LCMM-VAE (ours, $R{=}2$)} & $0.598_{\pm 0.028}$ & $0.668_{\pm 0.053}$ & \dash\rlap{$^{\dag}$} & \dash\rlap{$^{\dag}$} & 12.7 \\
  \bottomrule
  \end{tabular*}}
  \end{table*}

\section{Results}
\label{sec:results}

\subsection{Clustering}
\label{sec:res_clustering}

\paragraph{Simulation.}
On held-out ARI (\tableref{tab:main}), LCMM-VAE returns the best partition under both mechanisms ($0.90$ under MCAR, $0.76$ under MAR), significantly ahead of every baseline. The two-stage methods score at most $0.71$ under MCAR and $0.58$ under MAR, while no deep clustering baseline exceeds $0.24$. The random effects carry this lead: without them ($R{=}0$), ARI falls to $0.29$ and $0.27$. Removing the mixture costs less ($G{=}1$: $0.88$ and $0.69$), significantly only under MAR.

\begin{figure}[!b]
\floatconts
  {fig:gait_brace}
  {\caption{\textbf{Gait: LCMM-VAE class trajectories}. Solid: decoded class curve $f_\theta(B^{(g)}x_S)$; dashed: observed mean of the walks in each brace condition.}}
  {\centering\includegraphics[width=\linewidth]{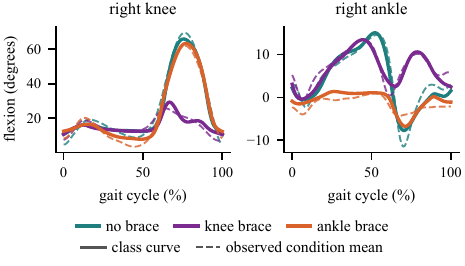}\par\vspace{-20pt}}
\vspace{-8pt}
\end{figure}

\paragraph{Gait.}
Gait is a control dataset, with a complete grid and classes that are easily recovered (\figureref{fig:gait_brace}). On complete curves, LCMM-VAE ($0.94$) is comparable to the three two-stage methods ($0.90$--$0.97$) and to its $G{=}1$ ablation ($1.00$). Missingness then separates the methods. Under \texttt{mcar}, every marker keeps observed points along the walk, to which the two-stage methods fit their splines: they keep their partition ($0.90$--$0.95$), and GMM-diag and FPCA $+$ $k$-means are significantly ahead of LCMM-VAE ($0.84$). Under \texttt{channel}, a removed marker has no observed point, so the two-stage methods fall ($0.11$--$0.45$). LCMM-VAE, which encodes each time point from all its markers, reaches $0.51$, comparable to IDEC ($0.52$) and CRLI ($0.46$) and significantly ahead of the other methods. Neither IDEC nor CRLI, however, returns class curves, whereas LCMM-VAE decodes each class in the markers' units (\figureref{fig:gait_brace}). Under both mechanisms, the mixture significantly outperforms its $G{=}1$ ablation.

\paragraph{Running time.}
One LCMM-VAE fit takes about $13$\,s on the simulation, less than any deep baseline we timed ($35$--$3{,}490$\,s), whereas the LCMM is infeasible at $K=50$ and its single gait fit did not finish within $6$\,h.

\subsection{Reconstruction}
\label{sec:res_recon}

\paragraph{Simulation.}
In MSE at hidden cells (\tableref{tab:recon}), linear interpolation is the most accurate method under both mechanisms ($0.557$ under MCAR, $0.617$ under MAR), followed by LCMM-VAE ($0.598$ and $0.668$). The simulated trajectories vary smoothly, so interpolating between a subject's neighbouring visits works well, but interpolation returns no partition. LCMM-VAE is significantly more accurate than GP-VAE and SAITS under both mechanisms and than BRITS under MAR, and comparable to BRITS under MCAR. Without random effects ($R{=}0$), MSE rises to $0.816$ and $0.949$.

\paragraph{Gait.}
Under \texttt{mcar}, where hidden values keep observed neighbours in time, linear interpolation, BRITS and SAITS are more than an order of magnitude more accurate than LCMM-VAE ($0.004$--$0.017$ against $0.277$), and LOCF and GP-VAE are also significantly better. Under \texttt{channel}, the methods with no learned model fall to the mean fill ($1.027$), and LCMM-VAE ($0.353$) outperforms every baseline, significantly ahead of GP-VAE ($0.470$), BRITS ($0.602$) and SAITS ($0.640$). With \texttt{channel} missingness, where the partition is at its weakest (ARI $0.51$, \tableref{tab:main}), LCMM-VAE's $G{=}1$ ablation is comparable ($0.338$). Under \texttt{mcar}, the mixture is significantly more accurate ($0.277$ against $0.371$). Structure within a walk thus wins when neighbours remain, and structure across walks when they do not.

\subsection{Clinical Data}
\label{sec:exp_physionet}
PhysioNet has no reference partition, so we primarily evaluate whether LCMM-VAE recovers clinically meaningful classes, assessed on variables the model never sees. The classes differ in patient profile: class D gathers the youngest patients, most often in surgical units, and class B most of the cardiac-surgery recovery stays, with the highest severity scores. They also differ in outcome, with in-hospital mortality ranging from $12.4\%$ (class A) to $18.7\%$ (class C), the class with the longest stays. These differences are significant on every seed. 

\paragraph{Further results.} \appendixref{app:extra} provides the paired tests behind every comparison, every method at every setting, sensitivity analyses to the latent dimension and the basis size, the selection of $G$, and, on PhysioNet, a more detailed description of the classes and a benchmark on mortality prediction. 

% =========================================================================
\section{Discussion}
\label{sec:discussion}
% =========================================================================
LCMM-VAE was designed for cohorts such as those collected by wearables and electronic health records. These combine many markers, sparse and irregular visits, and patients who depart from a typical course. On a simulation with these features, LCMM-VAE partitions the subjects significantly better than every baseline, while also describing each class by a trajectory in the units of the markers. Learning the classes jointly, rather than clustering a single-class fit afterwards, improved the partition in most incomplete settings. When a marker is missing for a whole recording, LCMM-VAE also reconstructs it better than every baseline. Where values keep close neighbours in time, methods that use them perform better. Structure within and across subjects thus appears complementary, which motivates combining the mixture with a scalable Gaussian-process prior that captures both \citep{Balik2025}.

\paragraph{Limitations.}
LCMM-VAE is built for real-world cohorts followed for years, with many markers recorded at irregular times and covariates beyond time. Such cohorts are rarely openly accessible, and the datasets considered here exercise only part of this design. Random effects are tested only on the simulation, whose data are drawn from the model's latent family. On gait, deviations sit at two nested levels, subject and cycle, whereas ours act at one. On PhysioNet, a $48$\,h window captures each subject's baseline more than its dynamics, so random intercepts would compete with the class trajectories. Applying the full model to a cohort of this kind, such as longitudinal wearable data, is the natural next step.

Independently of the data, three modelling limitations remain. First, time is an observed covariate, so every subject shares one clock, and the model cannot align processes that unfold at different paces, as individual time reparametrisations do \citep{Schiratti2017}. Second, the decoder is trained on codes inferred from each subject's data but reconstructs from the prior, a mismatch that the divergence term does not fully close \citep{Sohn2015}. Third, identifiability is proven with a fixed decoder and only argued with a learned one (\sectionref{sec:ident_main}); a full proof, in line with \citet{Kivva2022}, is left to future work.

\section{Conclusion}
\label{sec:conclusion}
We introduced the LCMM-VAE, which structures the latent space of a variational autoencoder as a mixture of class-specific linear mixed models, so that a class is a covariate-driven curve rather than a free point in a representation space. The construction is identifiable under stated conditions: up to a permutation of the class labels with a fixed decoder and up to a permutation and one affine map of the latent space with a learned piecewise-affine one. Its class responsibilities and mixed-model parameters are updated in closed form, leaving only the networks to gradient descent. On the data we study, it is the only method we compare to return a partition, a reconstruction and an interpretable trajectory per class from one model. Three directions follow: relaxing the parametric form of the class curves, for instance with a scalable Gaussian-process prior \citep{Balik2025}; giving each subject its own rate of progression; and modelling the observation process, so that data missing \emph{not} at random fall inside the same model.

\acks{This work was supported by Withings and Inria (Institut national de recherche en sciences et technologies du numérique) through the CIFRE programme of the ANRT (Association Nationale de la Recherche et de la Technologie).}

\clearpage

\bibliography{../v5/ref,refs_cr}

\clearpage
\appendix

% =========================================================================
\section{Notation}
\label{app:notation}

\tableref{tab:notation} lists the main symbols used throughout the paper. 

\begin{table}[H]
\floatconts
  {tab:notation}
  {\caption{\textbf{Main notation}.}}
  {\scriptsize\setlength{\tabcolsep}{3pt}
  \begin{tabular}{@{}l >{\raggedright\arraybackslash}p{0.60\linewidth}@{}}
\toprule
Symbol & Meaning \\
\midrule
$N$, $n_i$ & subjects; visits of subject $i$ \\
$K$, $L$, $S$, $R$, $G$ & markers, latent dim., $x_S$ dim., $x_R$ dim., classes \\
$y_{ij}\in\R^{K}$, $o_{ij}$ & observation at visit $j$; its observation mask \\
$z_{ij}\in\R^{L}$ & latent state at visit $j$ \\
$x_{S,ij}$, $x_{R,ij}$ & fixed- and random-effect design rows \\
$Y_i,Z_i,X_{S,i},X_{R,i}$ & the same, stacked by visit for subject $i$ \\
$c_i\in\{1,\dots,G\}$ & class of subject $i$ \\
$\pi$, $\pi_g$ & class prevalences, $\pi_g>0$, $\sum_g\pi_g=1$ \\
$A_S\in\R^{L\times S}$ & shared (population) coefficients \\
$\Delta^{(g)}$, $B^{(g)}$ & class-$g$ departure; $B^{(g)}=A_S+\Delta^{(g)}$ \\
$A_{R,i}\in\R^{L\times R}$ & subject-$i$ random effects, rows $a_{R,i,l}^{\top}$ \\
$D_l$, $D$ & covariance of $a_{R,i,l}$; $\blockdiag(D_1,\dots,D_L)$ \\
$\sigma_\varepsilon^{2}$, $\sigma_y^{2}$ & latent residual variance; marker noise \\
$f_\theta$, $g_\phi$ & decoder; encoder \\
$\mu_\phi,\sigma^2_\phi$ & encoder mean and variance heads \\
$\gamma_{ig}$ & variational class responsibility \\
$m_{i,l}$, $V_{i,l}$ & variational mean and covariance of $a_{R,i,l}$ \\
$\kappa_{ig}$ & cost of placing subject $i$ in class $g$, \eqref{eq:Kig} \\
$\Omega_{i,l}(D_l)$ & marginal covariance of latent coordinate $l$, \eqref{eq:Omega_block} \\
$\mathcal{L}_i$, $\beta$ & evidence lower bound; KL weight (warm-up in pretraining, $1$ after) \\
$\Theta$, $\psi$ & generative and variational parameters \\
\bottomrule
\end{tabular}}
\end{table}

% =========================================================================
\section{Derivation of the Evidence Lower Bound}
\label{app:elbo}
% =========================================================================

The joint density \eqref{eq:joint} rests on four assumptions:
\textbf{(A1)} $c_i$ and $A_{R,i}$ are independent of the covariate designs; \textbf{(A2)} $c_i\ind A_{R,i}$, and $c_i$ is time-invariant; \textbf{(A3)} given $(c_i,A_{R,i})$, the latent states are independent across visits; \textbf{(A4)} $y_{ij}$ depends on the other variables only through $z_{ij}$, so the decoder is pointwise and class-agnostic.

Write $\mathcal{Z}_i=(Z_i,A_{R,i},c_i)$ for the latent variables of subject $i$. Since $\log p_\Theta(Y_i\mid X_i)$ is free of $\mathcal{Z}_i$, it equals its own expectation under $q_\psi$; substituting Bayes' rule, $p_\Theta(Y_i\mid X_i)=p_\Theta(Y_i,\mathcal{Z}_i\mid X_i)/p_\Theta(\mathcal{Z}_i\mid Y_i,X_i)$, multiplying and dividing by $q_\psi(\mathcal{Z}_i\mid Y_i)$ and splitting the logarithm gives the exact identity
\begin{multline}\label{eq:elbo_def}
  \log p_\Theta(Y_i\mid X_i)
  = \mathcal{L}_i(\psi,\Theta)\\
  + \KL{q_\psi(\mathcal{Z}_i\mid Y_i)}{p_\Theta(\mathcal{Z}_i\mid Y_i,X_i)},
\end{multline}
with $\mathcal{L}_i:=\E_{q_\psi}\bigl[\log p_\Theta(Y_i,\mathcal{Z}_i\mid X_i)-\log q_\psi(\mathcal{Z}_i\mid Y_i)\bigr]$. The divergence is non-negative, so $\mathcal{L}_i$ is a lower bound on the evidence, tight when $q_\psi$ is the posterior. Factorising the joint as $p_\Theta(Y_i\mid\mathcal{Z}_i,X_i)\,p_\Theta(\mathcal{Z}_i\mid X_i)$ and applying \textbf{(A4)},
\begin{equation}\label{eq:lik_reduce}
  p_{\theta,\sigma_y^2}\bigl(Y_i\mid Z_i,A_{R,i},c_i,X_i\bigr)
  = p_{\theta,\sigma_y^2}\bigl(Y_i\mid Z_i\bigr),
\end{equation}
gives the familiar VAE form, reconstruction minus a divergence to the prior,
\begin{multline*}
  \mathcal{L}_i
  = \E_{q_\psi}\bigl[\log p_{\theta,\sigma_y^2}(Y_i\mid Z_i)\bigr]\\
  - \KL{q_\psi(\mathcal{Z}_i\mid Y_i)}{p_\Theta(\mathcal{Z}_i\mid X_i)}.
\end{multline*}
By \eqref{eq:lik_reduce}, all temporal structure and heterogeneity must pass through $Z_i$. Dropping $c_i$ is what makes the decoder class-agnostic, which is why term (i) does not appear in \eqref{eq:qc_iso}.

Substituting \eqref{eq:joint} and \eqref{eq:qfam}, the log-ratio inside the divergence splits by the latent variables each piece involves,
\begin{multline*}
  \log q_\psi - \log p_\Theta
  = \underbrace{\bigl[\log q_{\gamma_i}(c_i)-\log p_\pi(c_i)\bigr]}_{c_i\text{ only}}\\
  + \underbrace{\textstyle\sum_{l}\bigl[
      \log q_{m_{i,l},V_{i,l}}(a_{R,i,l})-\log p_{D_l}(a_{R,i,l})\bigr]}_{a_{R,i,l}\text{ only}}\\
  + \underbrace{\textstyle\sum_{j}\bigl[\log\qij-\log p^{(c_i)}_{ij}\bigr]}_{z_{ij},\,c_i,\,A_{R,i}} ,
\end{multline*}
with $p^{(g)}_{ij}:=\pijg$ the class-$g$ prior of visit $j$.
In expectation under $q_\psi$, the first group is $\KL{\gamma_i}{\pi}$, term (iii), and the second is $\KL{q(A_{R,i})}{p_D(A_{R,i})}$, term (iv). The third group involves all three latents, but \eqref{eq:qfam} makes $Z_i$, $A_{R,i}$ and $c_i$ independent, so we integrate them out one at a time. At fixed $c_i{=}g$ and $A_{R,i}$, the prior $p^{(g)}_{ij}$ is a fixed Gaussian and the codes are independent across visits, so the expectation over $Z_i$ is $\sum_j\KL{\qij}{\pijg}$. Its expectation over $q(A_{R,i})$ is $\kappa_{ig}$ \eqref{eq:Kig}, and the expectation over $c_i\sim\Cat(\gamma_i)$ is $\sum_g\gamma_{ig}\kappa_{ig}$, term (ii). This is \eqref{eq:elbo}. We now give each term in closed form, except (i).

\paragraph{Term (i): reconstruction.}
The decoder is a neural network, so the expectation has no closed form. We use the reparametrised estimator \citep{Kingma2014}, $z^{(t)}_{ij}=\mu_\phi(y_{ij})+\sigma_\phi(y_{ij})\odot\epsilon^{(t)}$ with $\epsilon^{(t)}\sim\Normal(\zerov{L},I_L)$. With \eqref{eq:lcmm_y} restricted to observed cells,
\begin{multline}\label{eq:imc}
  \widehat{\text{(i)}}=-\frac12\sum_{j=1}^{n_i}\sum_{k=1}^{K}o_{ij,k}
  \Bigg[\frac1T\sum_{t=1}^{T}
      \frac{\bigl(y_{ij,k}-f_{\theta,k}(z^{(t)}_{ij})\bigr)^{2}}
           {\sigma^{2}_{y,k}}\\
    + \log\bigl(2\pi\sigma^{2}_{y,k}\bigr)\Bigg],
\end{multline}
with $T=1$ throughout. The markers are conditionally independent given $z_{ij}$, so integrating out the hidden entries of $y_{ij}$ removes their factors: the bound is on the likelihood of the observed entries, and the mask $o_{ij,k}$ enters its formula only through this term \citep{Nazabal2020}. Hidden cells reach the encoder as zeros, which changes $q_\phi$ but not the validity of the bound.

\paragraph{Term (ii): class-weighted latent regularisation.}
Both $\qij$ and $\pijg$ are Gaussians with diagonal covariance, so their divergence is a sum over latent coordinates $l$. The $l$-th prior mean,
\begin{equation*}
  \bar z^{(g)}_{ij,l}
  = \bigl[B^{(g)}x_{S,ij}\bigr]_l + a_{R,i,l}^{\top}x_{R,ij},
\end{equation*}
depends on the random effects only through $a_{R,i,l}$, whose variational mean and covariance are $m_{i,l}$ and $V_{i,l}$. Taking the Gaussian divergence at a given $a_{R,i,l}$ and then its expectation, using $\E[X^2]=(\E X)^2+\mathrm{Var}(X)$ for $X=\mu_{\phi,ijl}-\bar z^{(g)}_{ij,l}$, gives
\begin{multline}\label{eq:kappa}
  \kappa^{(g)}_{ijl}
  = \frac{1}{2\sigma_\varepsilon^{2}}\Bigl[
      \underbrace{\bigl(\mu_{\phi,ijl}-\bar m^{(g)}_{ijl}\bigr)^{2}}_{\text{centre mismatch}}
      + \underbrace{x_{R,ij}^{\top}V_{i,l}\,x_{R,ij}}_{\text{uncertainty in }a_{R,i,l}}\Bigr]\\
  + \frac12\underbrace{\Bigl[\frac{\sigma^{2}_{\phi,ijl}}{\sigma_\varepsilon^{2}}-1
    -\log\frac{\sigma^{2}_{\phi,ijl}}{\sigma_\varepsilon^{2}}\Bigr]}_{\text{width mismatch}\;\ge\;0},
\end{multline}
where $\bar m^{(g)}_{ijl}=[B^{(g)}x_{S,ij}]_l+m_{i,l}^{\top}x_{R,ij}$ is the prior mean at $a_{R,i,l}=m_{i,l}$, and $\kappa_{ig}=\sum_j\sum_l\kappa^{(g)}_{ijl}$. The width term vanishes when the encoder's variance equals the prior's, $\sigma_\varepsilon^2=1$ here. Only the centre mismatch depends on $g$, so in \eqref{eq:gammastar} classes are compared by the squared distance from the encoder means to each class's prior mean; the other two terms change the value of the bound but cancel in the responsibilities.

\paragraph{Term (iii): class-prior regularisation.} A divergence between categorical distributions,
\begin{equation*}
  \KL{\gamma_i}{\pi}=\sum_{g=1}^{G}\gamma_{ig}\log\frac{\gamma_{ig}}{\pi_g}.
\end{equation*}

\paragraph{Term (iv): random-effect regularisation.}
Prior and variational covariances are block-diagonal in the same blocks, so the divergence is a sum of $L$ Gaussian divergences on $\R^{R}$,
\begin{multline}\label{eq:term_iv_closed}
  \KL{q(A_{R,i})}{p_{D}(A_{R,i})}
  =\frac12\sum_{l=1}^{L}\Bigl[\tr\bigl(D_l^{-1}V_{i,l}\bigr)\\
  + m_{i,l}^{\top}D_l^{-1}m_{i,l} - R
  - \log\det\bigl(D_l^{-1}V_{i,l}\bigr)\Bigr].
\end{multline}
The quadratic $m_{i,l}^{\top}D_l^{-1}m_{i,l}$ is the shrinkage penalty on the subject's deviation, identical in form to the one implicit in the Best Linear Unbiased Predictor (BLUP) of a linear mixed model; the remaining terms penalise a variational covariance that departs from the prior scale in either direction. At $R=0$ term (iv) is absent.

\paragraph{A structured alternative.}
The mean-field family
  \eqref{eq:qfam} drops the dependence between a subject's class and its deviation. A
  structured family $q(A_{R,i}\mid c_i)$, with one random-effect mean per class, would keep it; its responsibilities
  would then be the class-membership probabilities of the classical latent-class mixed model
  \citep{ProustLima2017}, computed on the encoder means. We leave it to future work.

% =========================================================================
\section{Training Procedure}
\label{app:mstep}

% =========================================================================

Every quantity of the mixed-model block maximises $\sum_i\mathcal{L}_i$ in closed form when the others, and the networks, are held fixed. Each derivation below collects the terms of \eqref{eq:elbo} that contain the quantity and solves for it in closed form. We write $r^{(g)}_{ijl}:=\mu_{\phi,ijl}-\bar m^{(g)}_{ijl}$ for the residual of the encoder mean against the class-$g$ trajectory, and $\bar m^{S}_{ijl}:=\sum_g\gamma_{ig}[B^{(g)}x_{S,ij}]_l$ for the responsibility-weighted class curve, so that
  \begin{equation*}
    \textstyle\sum_g\gamma_{ig}\bar m^{(g)}_{ijl}
    = \underbrace{\bar m^{S}_{ijl}}_{\text{class curves}}
    + \underbrace{m_{i,l}^{\top}x_{R,ij}}_{\text{random effect}} .
  \end{equation*}

\subsection{Closed-Form Updates}

\paragraph{The covariance $V_{i,l}$.}
$V_{i,l}$ enters term (ii) through $\sum_j x_{R,ij}^{\top}V_{i,l}x_{R,ij}$, identical in every class so that the weights $\gamma_{ig}$ sum out, and term (iv) through $\tr(D_l^{-1}V_{i,l})-\log\det V_{i,l}$. With $\partial(x^{\top}Vx)/\partial V=xx^{\top}$ and $\partial\log\det V/\partial V=V^{-1}$,
\begin{equation*}
  \frac{\partial\mathcal{L}_i}{\partial V_{i,l}}
  = -\frac{X_{R,i}^{\top}X_{R,i}}{2\sigma_\varepsilon^{2}}
    -\frac12 D_l^{-1}+\frac12 V_{i,l}^{-1}=0,
\end{equation*}
that is,
\begin{equation}\label{eq:Vstar}
  V_{i,l}^{-1}
  = \underbrace{D_l^{-1}}_{\text{prior precision}}
  + \underbrace{\tfrac{1}{\sigma_\varepsilon^{2}}
     X_{R,i}^{\top}X_{R,i}}_{\text{data precision}} .
\end{equation}
Both precisions push the covariance down, and the term $\log\det V_{i,l}$ of the bound, which tends to $-\infty$ as $V_{i,l}\to0$, prevents collapse. $V_{i,l}$ is positive definite by construction, so it needs no Cholesky  parametrisation, and it depends on the subject only through its visit schedule $\{x_{R,ij}\}_j$: more visits, spread more widely in time, give tighter random effects.

\paragraph{The random-effect mean $m_{i,l}$.}
Term (ii) contributes a fit term and term (iv) a shrinkage term,
\begin{multline}\label{eq:mgrad}
  \frac{\partial\mathcal{L}_i}{\partial m_{i,l}}
  = \underbrace{\frac{1}{\sigma_\varepsilon^{2}}\sum_{j}x_{R,ij}
      \Bigl[\sum_{g}\gamma_{ig}\,r^{(g)}_{ijl}\Bigr]}_{\text{fit, from (ii)}}\\
  \;-\;\underbrace{D_l^{-1}m_{i,l}}_{\text{shrinkage, from (iv)}} .
\end{multline}
With $x_{R,ij}=(1,t_{ij})^{\top}$, a subject lying consistently above its class curve raises its random intercept, and one drifting upwards raises its random slope. As the random-effect part of $\bar m^{(g)}_{ijl}$ is the same in every class, $\sum_g\gamma_{ig}r^{(g)}_{ijl}=\mu_{\phi,ijl}-\bar m^{S}_{ijl}-m_{i,l}^{\top}x_{R,ij}$. Setting \eqref{eq:mgrad} to zero and collecting the terms in $m_{i,l}$ gives
\begin{equation*}
  \Bigl[D_l^{-1}+\tfrac{1}{\sigma_\varepsilon^{2}}X_{R,i}^{\top}X_{R,i}\Bigr]m_{i,l}
  = \tfrac{1}{\sigma_\varepsilon^{2}}\sum_{j}x_{R,ij}
    \bigl(\mu_{\phi,ijl}-\bar m^{S}_{ijl}\bigr),
\end{equation*}
whose bracket is $V_{i,l}^{-1}$ by \eqref{eq:Vstar}, so
\begin{equation}\label{eq:mstar}
  m_{i,l}
  = \tfrac{1}{\sigma_\varepsilon^{2}}\,V_{i,l}
     \textstyle\sum_{j}x_{R,ij}\bigl(\mu_{\phi,ijl}-\bar m^{S}_{ijl}\bigr).
\end{equation}
Together, $(m_{i,l},V_{i,l})$ is the posterior of a Bayesian linear regression of the residuals $\mu_{\phi,ijl}-\bar m^{S}_{ijl}$ on $X_{R,i}$, with prior $\Normal(\zerov{R},D_l)$ and noise variance $\sigma_\varepsilon^2$. Its mean is a ridge regression per subject whose penalty is learned from the cohort, which is the BLUP of a linear mixed model \citep{LairdWare1982} with the encoder mean as response. 

\paragraph{The responsibilities $\gamma_i$.}
Only terms (ii) and (iii) contain $\gamma_i$: term (i) by \textbf{(A4)}, and term (iv) because neither $q(A_{R,i})$ (by \eqref{eq:qfam}) nor $p_D(A_{R,i})$ (by \textbf{(A2)}) involves $c_i$. Maximising $-\sum_g\gamma_{ig}\kappa_{ig}-\KL{\gamma_i}{\pi}$ on the simplex gives \eqref{eq:gammastar}; \appendixref{app:qc} gives the derivation and relates it to the mixture-prior VAEs. As $m_{i,l}$ depends on $\gamma_i$ through $\bar m^{S}_{ijl}$, and $\kappa_{ig}$ on $m_{i,l}$, the E-step updates them in turn; each update is exact given the other, so neither decreases the bound.

\paragraph{The covariances $D_l$ and prevalences $\pi$.}
$D_l$ appears only in term (iv). Rewriting \eqref{eq:term_iv_closed} with $m^{\top}D^{-1}m=\tr(D^{-1}mm^{\top})$, the part of $\sum_i\mathcal{L}_i$ that depends on $D_l$ is
\begin{equation}\label{eq:D_obj}
  -\tfrac12\tr\Bigl(D_l^{-1}\textstyle\sum_i\bigl(m_{i,l}m_{i,l}^{\top}+V_{i,l}\bigr)\Bigr)
  -\tfrac{N}{2}\log\det D_l .
\end{equation}
Using $\partial\tr(D^{-1}M)/\partial D=-D^{-1}MD^{-1}$ and $\partial\log\det D/\partial D=D^{-1}$, setting the derivative of \eqref{eq:D_obj} to zero and multiplying left and right by $D_l$ gives the cohort mean of the second moments $\E_q[a_{R,i,l}a_{R,i,l}^{\top}]=m_{i,l}m_{i,l}^{\top}+V_{i,l}$. The term $V_{i,l}$ matters: the means $m_{i,l}$ are shrunk towards zero, so averaging them alone would underestimate the spread of the random effects.

$\pi$ appears only in term (iii), so maximising $\sum_i\sum_g\gamma_{ig}\log\pi_g$ under $\sum_g\pi_g=1$ gives the average responsibility. Together,
  \begin{equation}\label{eq:pi_D_star}
  \begin{split}
    \pi_g &= \tfrac1N\textstyle\sum_i\gamma_{ig},\\
    D_l &= \tfrac1N\textstyle\sum_i\bigl(m_{i,l}m_{i,l}^{\top}+V_{i,l}\bigr).
  \end{split}
  \end{equation}

\paragraph{The class coefficients $B^{(g)}$.}
$A_S$ and $\Delta^{(g)}$ enter the bound only through $B^{(g)}=A_S+\Delta^{(g)}$, and for fixed $\pi$ the centring \eqref{eq:centering} maps any $\{B^{(g)}\}_g$ to exactly one admissible pair. We therefore maximise over the $G$ matrices $B^{(g)}$, which are unconstrained, and split afterwards. With $r^{(g)}_{ij}:=(r^{(g)}_{ijl})_l=\mu_\phi(y_{ij})-B^{(g)}x_{S,ij}-M_ix_{R,ij}\in\R^L$, the part of the bound that depends on $B^{(g)}$ is $-\tfrac{1}{2\sigma_\varepsilon^{2}}\sum_i\gamma_{ig}\sum_j\|r^{(g)}_{ij}\|^2$, whose gradient is
\begin{equation}\label{eq:ASgrad}
  \frac{\partial}{\partial B^{(g)}}\sum_{i}\mathcal{L}_i
  = \frac{1}{\sigma_\varepsilon^{2}}\sum_{i}\gamma_{ig}
    \sum_{j} r^{(g)}_{ij}x_{S,ij}^{\top}.
\end{equation}
Setting it to zero gives one weighted normal equation per class,
\begin{multline}\label{eq:Bstar}
  B^{(g)}\Bigl[\sum_{i}\gamma_{ig}\sum_{j}x_{S,ij}x_{S,ij}^{\top}\Bigr]\\
  =\sum_{i}\gamma_{ig}\sum_{j}\bigl(\mu_\phi(y_{ij})-M_ix_{R,ij}\bigr)x_{S,ij}^{\top},
\end{multline}
a least-squares regression of the encoder means, corrected for the random effects, on $x_S$, with each subject weighted by its responsibility for class $g$. The centring then fixes the split,
\begin{equation}\label{eq:ASrec}
  A_S=\sum_{g}\pi_g B^{(g)},\qquad \Delta^{(g)}=B^{(g)}-A_S,
\end{equation}
which satisfies $\sum_g\pi_g\Delta^{(g)}=\zerom{L}{S}$ by construction.

\subsection{Gradient-Trained Parameters}
The encoder $\phi$, the decoder $\theta$ and the noise variances $\sigma_y^2$ have no closed form, $\phi$ and $\theta$ being networks. $\sigma_y^2$ nearly has one: setting $\partial(\sum_i\mathcal{L}_i)/\partial\sigma^{2}_{y,k}=0$ gives
\begin{equation}\label{eq:sigmay}
  \sigma^{2}_{y,k}
  = \frac{\sum_{i,j}o_{ij,k}\,
      \E_{\qij}\bigl[(y_{ij,k}-f_{\theta,k}(z_{ij}))^{2}\bigr]}
     {\sum_{i,j}o_{ij,k}},
\end{equation}
the mean squared reconstruction error of marker $k$ over the visits where it was measured. The expectation still requires Monte Carlo through the decoder, so we leave the $K$ variances to the optimiser, with initial value $1$ and a softplus floor. Equation \eqref{eq:sigmay} explains the per-marker parametrisation: the weight $1/\sigma^{2}_{y,k}$ in \eqref{eq:imc} is the marker's own precision, so a poorly reconstructed marker is down-weighted rather than dominating the bound. 

\subsection{Training Algorithm}
\label{app:algo}
\label{app:training}
\figureref{fig:algo} assembles these updates; the encoder, the decoder and $\sigma_y^2$ take gradient steps. The E-step makes one sweep $V_{i,l}\to m_{i,l}\to\gamma_i$ per epoch, which is exact when $R=0$ but, when $R>0$, only moves towards the mean-field fixed point. The M-step is exact, since $\pi$, $D_l$ and $B^{(g)}$ appear in disjoint terms. The closed-form block thus never decreases the bound, whereas the finite stochastic gradient steps can: the scheme is a variational EM in the partial-step sense of \citet{NealHinton1998}, monotone within the closed-form block but not across epochs. The block needs every visit of every subject, so it runs on one encoder pass without gradient; only the gradient epoch is minibatched. After the last epoch, one more E-step and M-step align the prior with the final encoder.

\begin{figure}[t]
\floatconts
  {fig:algo}
  {\caption{\textbf{Training and inference algorithm}.}}
  {\begin{inlinealgo}
  \footnotesize
  \DontPrintSemicolon
  \KwIn{visits $y_{ij}$, designs $x_{S,ij},x_{R,ij}$, num.\ of classes $G$, stopping threshold $\delta$, max.\ epochs $E$}
  \textbf{Initialise:} pretrain a linear-prior VAE $\to\phi,\theta,\sigma_y^2$; initialise $\pi,\{B^{(g)}\},D$ on the encoder means \;
  \For{\textrm{epoch} $e=1,\dots,E$}{
    \tcp{E-step, closed form}
    \For{\textrm{each subject} $i$}{
      $V_{i,l}$ \eqref{eq:Vstar};\; $m_{i,l}$ \eqref{eq:mstar};\; $\gamma_{ig}$ \eqref{eq:gammastar};\; $\hat c_i=\arg\max_g\gamma_{ig}$\;
    }
    \tcp{M-step, closed form}
    $\pi, D_l$ \eqref{eq:pi_D_star};\; $B^{(g)}$ \eqref{eq:Bstar}, split into $A_S,\Delta^{(g)}$ by \eqref{eq:centering};\; recompute $V_{i,l}$, $m_{i,l}$\;
    \tcp{gradient}
    gradient steps on $\phi,\theta,\sigma_y^2$ (one epoch)\;
    \lIf{$e>1$ \textrm{and the fraction of subjects whose} $\hat c_i$ \textrm{changed is} $<\delta$}{\textbf{stop}}
  }
  E-step and M-step with the final encoder\;
  \tcp{inference, any subject}
  E-step on its observed visits, parameters fixed; decode by \eqref{eq:hybrid}\;
  \KwOut{responsibilities $\gamma_{ig}$ and classes $\hat c_i$; class trajectories $f_\theta(B^{(g)}x_S)$; for any subject, its reconstruction \eqref{eq:hybrid} at any time}
  \end{inlinealgo}}
\end{figure}

% =========================================================================
\section{Optimal Class Responsibilities}
\label{app:qc}
% =========================================================================
This appendix derives the optimal class responsibilities \eqref{eq:gammastar}, $\gamma_{ig}\propto\pi_g\exp(-\kappa_{ig})$: the variational probability that subject $i$ belongs to class $g$, where $\kappa_{ig}$ \eqref{eq:Kig} is the expected divergence between the subject's encoded visits and the prior of class $g$, summed over its visits. We derive it directly from the bound, then show that it is the optimum that mixture-prior VAEs express through the exact class posterior.

\paragraph{Direct derivation.} Only terms (ii) and (iii) of \eqref{eq:elbo} contain $\gamma_i$, and maximising
\begin{equation}\label{eq:qc_iso}
  \mathcal{L}^{\gamma}_i(\gamma_i)=
  -\sum_{g}\gamma_{ig}\kappa_{ig}
  -\sum_{g}\gamma_{ig}\log\frac{\gamma_{ig}}{\pi_g}
\end{equation}
under $\sum_g\gamma_{ig}=1$ with a Lagrange multiplier $\lambda$ gives the stationarity condition $-\kappa_{ig}-\log\gamma_{ig}-1+\log\pi_g+\lambda=0$, hence $\gamma_{ig}\propto\pi_g\exp(-\kappa_{ig})$. Term (i) is free of $\gamma_i$ because the decoder is class-agnostic by \textbf{(A4)}, and term (iv) because neither $q(A_{R,i})$ (by \eqref{eq:qfam}) nor $p_D(A_{R,i})$ (by \textbf{(A2)}) involves $c_i$. The entropy is strictly concave and the remainder linear, so this is the unique maximiser on the simplex: a softmax of $\log\pi_g-\kappa_{ig}$, the prior prevalence tilted by how badly class $g$ explains the encoder output, aggregated over \emph{all} of the subject's visits.

This appendix reconciles that with the form used by \citet{Jiang2017}, \citet{Falck2021} and \citet{Courrier2026}, which is stated as a divergence against the exact class posterior $p(c_i\mid Z_i,A_{R,i},X_i)$, a quantity \eqref{eq:elbo} does not display explicitly. Below, $\E_i$ denotes expectation under $\qZi\,\qAi$.

\paragraph{Regrouping the bound.}
To compare with these methods, we rewrite the class terms of the bound, $\text{(ii)}+\text{(iii)}=-\mathcal{L}^{\gamma}_i$ (terms (ii) and (iii) of \eqref{eq:elbo} without their minus signs), so that the exact class posterior appears. By \textbf{(A3)} the prior factorises over visits at fixed $(c_i,A_{R,i})$, and by \eqref{eq:qfam} so does $q_\phi$, so \eqref{eq:Kig} collects into
\begin{equation}\label{eq:kappa_agg}
  \kappa_{ig}
  = \E_i\bigl[\log \qZi - \log \pZig\bigr].
\end{equation}
By \textbf{(A1)} and \textbf{(A2)}, $p(c_i{=}g\mid A_{R,i},X_i)=\pi_g$, so Bayes' rule against the class-marginal prior $p(Z_i\mid A_{R,i},X_i)=\sum_{g'}\pi_{g'}\,p(Z_i\mid c_i{=}g',A_{R,i},X_i)$ gives
\begin{multline}\label{eq:bayes_c}
  \log \pZig\\
  = \log p(c_i{=}g\mid Z_i,A_{R,i},X_i)\\
  + \log p(Z_i\mid A_{R,i},X_i) - \log\pi_g .
\end{multline}
Substituting into \eqref{eq:kappa_agg}, weighting by $\gamma_{ig}$, summing and adding term (iii), the $\pm\log\pi_g$ cancel and the two $g$-free expectations leave the sum because $\sum_g\gamma_{ig}=1$:
\begin{multline}\label{eq:agg}
  \text{(ii)}+\text{(iii)}\\
  = \E_{\qAi}\Bigl[\KL{\qZi}{p(Z_i\mid A_{R,i},X_i)}\Bigr]\\
  + \E_i\Bigl[\KL{\gamma_i}{p(c_i\mid Z_i,A_{R,i},X_i)}\Bigr] .
\end{multline}
Only the second summand carries $\gamma_i$, so maximising the bound over the class factor minimises it. \eqref{eq:agg} is exactly $-\mathcal{L}^{\gamma}_i$ \eqref{eq:qc_iso}.

\paragraph{Why the divergence cannot be driven to zero.}
\citet{Jiang2017} set the second summand of \eqref{eq:agg} to zero, concluding $\gamma_i=p(c_i\mid Z_i,A_{R,i},X_i)$ and, on taking expectations, the arithmetic mean $\gamma_i=\E_i[p(c_i\mid Z_i,A_{R,i},X_i)]$. As \citet{Falck2021} observe, reaching zero would require that equality to hold simultaneously for \emph{every} $(Z_i,A_{R,i})$ in the support of $\qZi\,\qAi$, which is impossible unless the posterior does not vary with them: the left-hand side is a single distribution, the same for all $(Z_i,A_{R,i})$, while the right-hand side changes with them. Zero divergence is therefore generally unattainable, and the minimiser is not any individual $p(c_i\mid Z_i,A_{R,i},X_i)$ but the single distribution closest to all of them on average, a normalised \emph{geometric} rather than arithmetic mean,
\begin{equation}\label{eq:geom}
  \gamma_{ig}\;\propto\;
  \exp\Bigl(\E_i\bigl[\log p(c_i{=}g\mid Z_i,A_{R,i},X_i)\bigr]\Bigr).
\end{equation}

\paragraph{Closed-form evaluation.}
\citet{Falck2021} and \citet{Courrier2026} evaluate the expectation in \eqref{eq:geom} by drawing $T$ Monte Carlo samples $Z_i^{(t)}\sim q_\phi(Z_i\mid Y_i)$. Here it is exact. Taking $\E_i$ of \eqref{eq:bayes_c} splits it into three pieces:
\begin{multline}\label{eq:eta_exact}
  \E_i\bigl[\log p(c_i{=}g\mid Z_i,A_{R,i},X_i)\bigr]
  = \log\pi_g\\
  + \E_i\bigl[\log \pZig\bigr]\\
  - \underbrace{\E_i\bigl[\log p(Z_i\mid A_{R,i},X_i)\bigr]}
      _{\text{intractable, free of }g} .
\end{multline}
The last piece is the expected logarithm of a mixture, which has no closed form under $q_\phi$, but it does not depend on $g$. The middle piece follows from \eqref{eq:kappa_agg}: since $\E_i[\log\qZi]=-\sum_j\Ent(q_\phi(z_{ij}\mid y_{ij}))$, with $\Ent$ the differential entropy,
\begin{multline}\label{eq:eta_kl}
  \E_i\bigl[\log \pZig\bigr]\\
  = -\kappa_{ig} - \underbrace{\textstyle\sum_{j}\Ent\bigl(q_\phi(z_{ij}\mid y_{ij})\bigr)}_{\text{free of }g},
\end{multline}
where $\kappa_{ig}=\sum_j\sum_l\kappa^{(g)}_{ijl}$ is the closed form \eqref{eq:kappa}, and the entropy is free of $g$ because $q_\phi$ does not see the class. Together, \eqref{eq:eta_exact} and \eqref{eq:eta_kl} give
\begin{equation*}
  \E_i\bigl[\log p(c_i{=}g\mid Z_i,A_{R,i},X_i)\bigr]
  = \log\pi_g - \kappa_{ig} + C_i,
\end{equation*}
where $C_i$ collects the two $g$-free terms,
\begin{equation*}
  C_i := -\textstyle\sum_{j}\Ent\bigl(q_\phi(z_{ij}\mid y_{ij})\bigr)
  - \E_i\bigl[\log p(Z_i\mid A_{R,i},X_i)\bigr].
\end{equation*}
A softmax is invariant to a shift free of $g$, hence
\begin{equation*}
\begin{split}
  \gamma_{ig}
  &= \operatorname*{softmax}_{g}\bigl(\log\pi_g-\kappa_{ig}+C_i\bigr)\\
  &= \operatorname*{softmax}_{g}\bigl(\log\pi_g-\kappa_{ig}\bigr),
\end{split}
\end{equation*}
recovering \eqref{eq:gammastar}. Only $\kappa_{ig}$ survives, and it is in closed form.

% =========================================================================

% =========================================================================
\section{Identifiability}
\label{app:identifiability}
\label{app:ident}
% =========================================================================

\subsection{The Marginal Law in Latent Space}
Identify $Z_i$ with its \emph{coordinate-major} vectorisation
\begin{equation}\label{eq:Zi_vec}
  \vecop(Z_i^\top)
  = \bigl(Z_i^{(1)\top},\dots,Z_i^{(L)\top}\bigr)^\top\in\R^{Ln_i},
\end{equation}
$Z_i^{(l)}=(z_{i1,l},\dots,z_{in_i,l})^\top$, so the $l$-th block collects latent coordinate $l$ across all of subject $i$'s visits. Coordinate $l$ is then a univariate linear mixed model,
\begin{equation}\label{eq:block_lmm}
  Z_i^{(l)}=X_{S,i}\,b^{(g)}_l+X_{R,i}\,a_{R,i,l}+\varepsilon^{(l)}_i,
\end{equation}
with $b^{(g)\top}_l$ the $l$-th row of $B^{(g)}=A_S+\Delta^{(g)}$. Marginalising $A_{R,i}$ out of \eqref{eq:block_lmm} gives, conditionally on $c_i=g$,
\begin{equation}\label{eq:Zi_mixture}
  \vecop(Z_i^\top)\mid c_i{=}g \sim
  \Normal\Bigl(\bigl(I_L\otimes X_{S,i}\bigr)\vecop\bigl(B^{(g)\top}\bigr),
    \Omega_i(D)\Bigr),
\end{equation}
\begin{equation}\label{eq:Omega_block}
  \Omega_i(D)
  = \blockdiag_{l=1}^{L}\bigl(X_{R,i}D_lX_{R,i}^{\top}\bigr)
    + \sigma_\varepsilon^{2}I_{Ln_i},
\end{equation}
using $(I_L\otimes X_{R,i})D(I_L\otimes X_{R,i})^{\top}=\blockdiag_l(X_{R,i}D_lX_{R,i}^{\top})$. Two features of \eqref{eq:Omega_block} do the work below. It does not depend on $g$: neither the random effects nor the residual vary across classes. And it is exactly block-diagonal in $l$, because the $a_{R,i,l}$ are independent across coordinates and the residual $\sigma_\varepsilon^2I_L$ is isotropic, so it contributes $\sigma_\varepsilon^2I_{Ln_i}$ with no cross-coordinate correlation; the $l$-th block, $\Omega_{i,l}(D_l)$, is the covariance of a standard univariate linear mixed model. Marginalising the class as well,
\begin{multline}\label{eq:pZ_mixture}
  P_{Z_i}:\quad p\bigl(\vecop(Z_i^\top)\bigr)=\sum_{g=1}^{G}\pi_g\\
    \times\Normal\Bigl(\bigl(I_L\otimes X_{S,i}\bigr)\vecop\bigl(B^{(g)\top}\bigr),
      \Omega_i(D)\Bigr),
\end{multline}
a genuine $G$-component mixture: by \eqref{eq:centering} its \emph{mean} is $(I_L\otimes X_{S,i})\vecop(A_S^\top)$, but the components do not collapse at a subject whose design separates the class means, such as $i_0$ below, and its variance exceeds $\Omega_i(D)$ by the between-class term. Setting $G=1$ recovers the latent model of \citet{Ong2025}; taking $f_\theta$ known recovers that of \citet{ProustLima2017}.

\subsection{Identifiability of the Latent LCMM}
We recall that $\sigma_\varepsilon^{2}$ has been fixed to $1$.
\begin{lemma}\label{lem:lcmm}
Assume:
\begin{enumerate}\itemsep0pt
    \renewcommand{\theenumi}{C\arabic{enumi}}%
    \renewcommand{\labelenumi}{(\theenumi)}%
    \item\label{cond:re} the family $\{X_{R,i}\}_{i=1}^N$ is such that $D_l\mapsto\{X_{R,i}D_lX_{R,i}^{\top}\}_{i=1}^N$ is injective on symmetric $R\times R$ matrices;
    \item\label{cond:pi} $\pi_g>0$ for every $g$, and the $\pi_g$ are pairwise distinct;
    \item\label{cond:design} the $\Delta^{(g)}$ are pairwise distinct, and there exists a subject $i_0$ whose fixed-effect design $X_{S,i_0}$ has full column rank $S$;
    \item\label{cond:center} $\sum_{g}\pi_g\Delta^{(g)}=\zerom{L}{S}$.
\end{enumerate}
Then, within the parameter sets satisfying Conditions~\ref{cond:pi}--\ref{cond:center}, $\bigl(A_S,\{\Delta^{(g)}\}_{g},D,\{\pi_g\}_{g}\bigr)$ is identifiable from the collection of laws $P_Z=(P_{Z_i})_{i=1}^N$ of \eqref{eq:pZ_mixture}, up to a permutation of the class labels.
\end{lemma}

\begin{proof}
Suppose
\[(A_S,\{\Delta^{(g)}\},D,\{\pi_g\})\quad\text{and}\quad(\widetilde A_S,\{\widetilde\Delta^{(g)}\},\widetilde D,\{\widetilde\pi_g\})\]
induce $p(\vecop(Z_i^\top))=\tilde p(\vecop(Z_i^\top))$ for every $i$.

\emph{Subject $i_0$ has $G$ pairwise distinct components.} $X_{S,i_0}$ has full column rank, hence so does $I_L\otimes X_{S,i_0}$, so for $g\ne g'$ the equality of the two means forces $B^{(g)}=B^{(g')}$, that is $\Delta^{(g)}=\Delta^{(g')}$, contradicting Condition~\ref{cond:design}. The covariance $\Omega_{i_0}(D)$ being common to all classes, the $G$ components of subject $i_0$'s mixture are therefore pairwise distinct.

\emph{Labels.} By classical identifiability of finite Gaussian mixtures \citep{Teicher1963,YakowitzSpragins1968} applied at $i_0$, there is a permutation $\sigma$ with $\pi_g=\widetilde\pi_{\sigma(g)}$ for every $g$. Since $\{\pi_g\}$ does not depend on $i$ and is pairwise distinct (Condition~\ref{cond:pi}), this same $\sigma$ matches the two value sets at every subject; relabel so that $\sigma$ is the identity, the result holding up to a permutation of the class labels. Then, for every $g$ and every $i$,
\[
  \bigl(I_L\otimes X_{S,i}\bigr)\vecop\bigl(B^{(g)\top}\bigr)
  = \bigl(I_L\otimes X_{S,i}\bigr)\vecop\bigl(\widetilde B^{(g)\top}\bigr),
\]
and $\Omega_i(D)=\Omega_i(\widetilde D)$.

\emph{Means.} Stacking the first equality over every subject gives $(I_L\otimes X_S)\vecop(B^{(g)\top}-\widetilde B^{(g)\top})=0$ with $X_S$ the pooled design. Since $X_{S,i_0}$ is a row-submatrix of $X_S$, Condition~\ref{cond:design} implies $X_S$ has full column rank $S$, hence $I_L\otimes X_S$ does, hence $B^{(g)}=\widetilde B^{(g)}$ for every $g$. By Condition~\ref{cond:center}, $A_S=\sum_g\pi_gB^{(g)}=\sum_g\widetilde\pi_g\widetilde B^{(g)}=\widetilde A_S$, and therefore $\Delta^{(g)}=\widetilde\Delta^{(g)}$.

\emph{Variance.} $\Omega_i(D)=\Omega_i(\widetilde D)$ is an equality of two block-diagonal matrices with identical block structure; subtracting the common $\sigma_\varepsilon^{2}I_{Ln_i}$ leaves $X_{R,i}D_lX_{R,i}^{\top}=X_{R,i}\widetilde D_lX_{R,i}^{\top}$ for every $l$ and every $i$, and Condition~\ref{cond:re} gives $D_l=\widetilde D_l$, hence $D=\widetilde D$.
\end{proof}

Condition~\ref{cond:design} plays two roles. It subsumes the usual full-rank requirement on the pooled design, classical in the mixed-model literature and what makes the $B^{(g)}$ identifiable; and it guarantees that subject $i_0$'s mixture has $G$ pairwise distinct components, so that the class labelling can be read off at that subject. It is the fixed-effect analogue of a condition used for random-effect designs in \citet{Demidenko2013} (see also \citealp{Amestoy2024}), where identifiability likewise follows from the existence of a single individual whose design matrix has full column rank. It is only sufficient: the minimal requirement is that the $G$ component means be pairwise distinct at some subject, which involves the unknown $B^{(g)}$ as well as the design, whereas the full-rank formulation can be checked from the design alone. Condition~\ref{cond:re}, by contrast, is necessary and sufficient for identifiability of the random-effect covariance \citep{Amestoy2024}.

\begin{remark}[Basis size]\label{rem:nbasis}
Condition~\ref{cond:design} requires $n_{i_0}\ge S$: at least one subject must have been seen at least as many times as $x_{S,i_0 j}$ has columns. Since $S$ is the number of basis functions plus the number of time-invariant covariates, the basis cannot be refined indefinitely --- a cohort whose best-observed subject has ten visits admits at most ten fixed-effect columns if the lemma is to apply, whatever the total number of observations $N_\bullet$. This is a condition on the \emph{best-observed} subject, not on all of them, so it is compatible with a cohort in which most subjects contribute a handful of visits. It is also why $\{X_{R,i}\}$ in Condition~\ref{cond:re} is required as a family rather than per subject: the variance component pools across the cohort while the labelling does not.
\end{remark}

\begin{remark}[Block-diagonal $D$]\label{rem:block}
Because $x_{R,ij}$ carries no coordinate index, the same design $\{X_{R,i}\}_i$ appears in all $L$ blocks. Condition~\ref{cond:re} therefore need only be verified \emph{once}, and it then yields $D_l=\widetilde D_l$ for all $l$ simultaneously: identifiability of an $LR\times LR$ matrix becomes $L$ copies of an already-solved $R\times R$ problem, with no strengthening of the design condition. The diagonality of the latent residual covariance (here isotropic) is what makes the decoupling exact; a non-diagonal $\Sigma_\varepsilon$ would couple coordinates within a visit and the blocks would no longer separate. The residual variance is held fixed above, which is what the proof needs when subtracting the common $\sigma_\varepsilon^2I_{n_i}$; following Proposition 2 of \citet{Amestoy2024} it could instead be treated as a second random effect with design $I_{n_i}$ and identified alongside $D_l$.
\end{remark}

\subsection{From the Latent Space to the Observations: Fixed Decoder}

\begin{assumption}\label{ass:decoder}
For every subject $i$ and any two admissible laws $P_{Z_i},\widetilde P_{Z_i}$ of the stacked latent trajectory,
\begin{multline*}
\int p_{\theta,\sigma_y^2}(Y_i\mid Z_i)\,dP_{Z_i}\\
= \int p_{\theta,\sigma_y^2}(Y_i\mid Z_i)\,d\widetilde P_{Z_i}
\ \text{ for a.e.\ }Y_i
\end{multline*}
implies $P_{Z_i}=\widetilde P_{Z_i}$.
\end{assumption}

\noindent Assumption~\ref{ass:decoder} is not primitive: it follows from injectivity of the decoder.

\begin{lemma}\label{lem:decoder}
If the per-visit decoder $f_\theta:\R^L\to\R^K$ is continuous and injective, which requires
$K\ge L$, then Assumption~\ref{ass:decoder} holds.
\end{lemma}

\begin{proof}
The visit-wise map
\[F_\theta(Z_i)=\bigl(f_\theta(z_{i1}),\dots,f_\theta(z_{in_i})\bigr)\]
is continuous
and injective, so by the Lusin--Souslin theorem the pushforward
$P_{Z_i}\mapsto(F_\theta)_\#P_{Z_i}$ is injective on Borel measures. The observation law is
the convolution
$P_{Y_i}=(F_\theta)_\#P_{Z_i}*\Normal(0,I_{n_i}\otimes\diag(\sigma_y^2))$, and a Gaussian
characteristic function never vanishes, so convolution with it is injective as well
\citep{Khemakhem2020}. Composing the two injections gives the claim. 
\end{proof}

Injectivity of $f_\theta$ is a property of the decoder alone, not of the noise model. \citet{Kivva2022} show that LeakyReLU networks whose layer widths never decrease from the latent space to the output are generically injective (their Cor.~H.4). \citet{Khemakhem2020ICEBeeM} use the same family of explicit MLP architectures (LeakyReLU activations, full-rank weights, monotonically ordered layer widths).

\begin{proposition}\label{prop:obs}
Under Assumption~\ref{ass:decoder} and the conditions of Lemma~\ref{lem:lcmm}, and with the decoder $f_\theta$ and the observation variances $\sigma_y^{2}$ held fixed and common to both parameter sets, the parameters $\bigl(A_S,\{\Delta^{(g)}\}_{g},D,\{\pi_g\}_{g}\bigr)$ are identifiable from the observed-data distribution of $(Y_{ij})$, up to a permutation of the latent-class labels. 
\end{proposition}
\begin{proof}
Throughout, $\theta$ and $\sigma_y^{2}$ are held fixed and common to both parameter sets;
only the latent-side parameters vary. By Assumption~\ref{ass:decoder},
$p_{\theta,\sigma_y^2}(Y_i)=\tilde p_{\theta,\sigma_y^2}(Y_i)$ implies $P_{Z_i}=\widetilde P_{Z_i}$ for every $i$,
that is $P_Z=\widetilde P_Z$ in the sense of Lemma~\ref{lem:lcmm}, which then gives the
parameter equalities up to a permutation of the class labels.
\end{proof}

\subsection{Learned Decoder}
\label{app:ident_learned}
\begin{remark}[Learned decoder]\label{rem:learned}
Proposition~\ref{prop:obs} holds $f_\theta$ and $\sigma_y^2$ fixed. We expect it to extend to a learned decoder, at the price of one affine map of the latent space. We sketch the argument here and leave its full proof to future work. \citet[Cor.~3.1 and Thm.~3.2(a,c)]{Kivva2022} show that, with a continuous, injective and piecewise-affine decoder and noise of known distribution, a Gaussian-mixture latent and the decoder are identified up to an invertible affine map. Stacking a subject's visits gives their setting: by \eqref{eq:pZ_mixture} the stacked latent state is a Gaussian mixture with at most $G$ components, and the stacked decoder applies $f_\theta$ visit by visit. Since $f_\theta$ is injective and shared across visits and subjects, the affine map should act visit by visit, through one map $h$ of $\R^L$ common to every visit and subject. With Condition~\ref{cond:design} and a design containing an intercept, the model would then be identified up to a class permutation and $h$. $A_S$, the $\Delta^{(g)}$ and $D$ would be determined only up to $h$, whereas the proportions, the partition and the decoded class trajectories would be unchanged. The stronger result of \citet[Thm.~3.2(b)]{Kivva2022}, up to permutation, scaling and translation, needs a condition (P2) that fails here: the classes share one latent covariance, so the ratios of their variances all equal one. The argument also needs a known $\sigma_y^2$, a piecewise-affine decoder, which our ELU decoder is not, and an injective decoder. Injectivity holds generically for LeakyReLU networks whose widths never decrease from the latent space to the output \citep[Cor.~H.4]{Kivva2022}.
\end{remark}

\begin{remark}[What the conditions cost]\label{rem:cost}
\emph{Condition~\ref{cond:pi} excludes balanced mixtures.} Distinctness of the $\pi_g$ is used only to make the permutation $\sigma$ independent of $i$, and it is a real restriction: it rules out $\pi=(1/G,\dots,1/G)$, the design of our own gait control, on which LCMM-VAE nonetheless recovers most of the partition (\tableref{tab:main}). Two routes remove it. If the design is random with a distribution whose support is connected, $g\mapsto B^{(g)}x$ is continuous in $x$ and the labelling can be propagated by continuity from any single separating design row. Alternatively, if the labelling is fixed at $i_0$ by distinctness of that subject's component means alone (which Condition~\ref{cond:design} already gives), and every other subject shares at least one design row with $i_0$, the labelling propagates without using $\pi$ at all. We state the lemma with the stronger condition because it is the version we can prove in a few lines; a reader whose classes are balanced should use one of the two alternatives.

\emph{The decoder's identifiability.} Assumption~\ref{ass:decoder} asks that the observation model be identifiable with respect to the latent distribution, and Lemma~\ref{lem:lcmm} then does the work on the latent side. The division is deliberate, since it isolates what the latent-class mixed-effects structure contributes, but it treats $f_\theta$ and $\sigma_y^2$ as known. Remark~\ref{rem:learned} sketches how the decoder could be identified as well, up to one affine map of the latent space. For our fitted model (ELU decoder, learned $\sigma_y^2$), Proposition~\ref{prop:obs} applies only conditionally, treating its fitted decoder and noise variances as known and that decoder as injective, which our architecture does not guarantee.
\end{remark}

\subsection{Empirical Check of the Conditions}
\label{app:decoder_families}
Remark~\ref{rem:learned} assumes a continuous, injective and piecewise-affine decoder and a known $\sigma_y^2$, whereas our decoder uses ELU activations, learns $\sigma_y^2$ and, on the simulation, narrows at the output ($3\to128\to256\to50$), so the injectivity result of \citet[Cor.~H.4]{Kivva2022} does not apply. We therefore refit LCMM-VAE on the simulation with only the decoder's activation replaced, by tanh or by one of two piecewise-affine activations, LeakyReLU and a leaky hard-tanh, and with equal class proportions ($1/3$ each, against $0.5/0.3/0.2$ in \appendixref{app:exp_sim}), which Condition~\ref{cond:pi} excludes. Besides ARI, we report the mean correlation coefficient (MCC) between the learned and true latents after a least-squares affine alignment \citep{Khemakhem2020ICEBeeM,Kivva2022}. Across activations and proportions (\tableref{tab:decoders}), ARI lies within $0.90$--$0.92$ under MCAR and $0.76$--$0.81$ under MAR, and MCC within $0.84$--$0.87$ and $0.81$--$0.84$. The ARI spreads are well below LCMM-VAE's seed-to-seed standard deviations ($0.08$ and $0.20$, \tableref{tab:main}): on this simulation, neither metric shows a detectable dependence on the activation or on equal proportions.

\begin{table}[H]
\centering
\caption{\textbf{Simulation: decoder families}, ARI and MCC, $N=500$, mean over seeds $0$--$4$.}
\label{tab:decoders}
\small\setlength{\tabcolsep}{5pt}
\begin{tabular}{@{}l cc cc@{}}
\toprule
& \multicolumn{2}{c}{ARI} & \multicolumn{2}{c}{MCC} \\
\cmidrule(lr){2-3}\cmidrule(lr){4-5}
Decoder & MCAR & MAR & MCAR & MAR \\
\midrule
\multicolumn{5}{@{}l}{\itshape Unequal proportions ($0.5/0.3/0.2$)} \\
ELU (ours) & $0.90$ & $0.76$ & $0.86$ & $0.83$ \\
tanh & $0.92$ & $0.81$ & $0.87$ & $0.84$ \\
LeakyReLU & $0.92$ & $0.77$ & $0.84$ & $0.83$ \\
leaky hard-tanh & $0.91$ & $0.80$ & $0.87$ & $0.84$ \\
\midrule
\multicolumn{5}{@{}l}{\itshape Equal proportions ($1/3$ each)} \\
ELU (ours) & $0.90$ & $0.78$ & $0.86$ & $0.81$ \\
tanh & $0.91$ & $0.77$ & $0.87$ & $0.82$ \\
LeakyReLU & $0.92$ & $0.79$ & $0.85$ & $0.81$ \\
leaky hard-tanh & $0.91$ & $0.76$ & $0.87$ & $0.81$ \\
\bottomrule
\end{tabular}
\end{table}

% =========================================================================
\section{Relation to the LMM-VAE (\texorpdfstring{$G{=}1$}{G=1})}
\label{app:lmmvae}
% =========================================================================

LCMM-VAE builds on the LMM-VAE of \citet{Ong2025}, which places a linear mixed model in the latent space of a VAE. We add a latent-class level, learn the random-effect covariances, and update the mixed-model block in closed form. Setting $G=1$ deletes $c_i$, $\pi$ and $\{\Delta^{(g)}\}$ and reduces \eqref{eq:elbo} to a VAE with prior mean $A_Sx_{S,ij}+A_{R,i}x_{R,ij}$: the LMM-VAE, up to two differences. They fix $a_{R,i,l}\sim\Normal(0,I_R)$ where we learn $D_l$ by \eqref{eq:pi_D_star}; and their two variants parametrise the subject level differently: the \emph{structured} sLMM-VAE shares one residual per subject (a static latent shift), while the \emph{ordinary} oLMM-VAE carries subject identity through one-hot columns of $A_R$, richer but with a design that grows in $N$. Our $x_{R,ij}$ is a small per-visit design, typically $(1,t_{ij})$, so the parameter count does not grow with the cohort. As an ablation it differs from their model in training too: it shares LCMM-VAE's pretraining and closed-form updates, runs for as many epochs as LCMM-VAE ran on the same seed, and is partitioned by the same em-EM procedure, so that only the number of classes differs.

\citet{Ong2025} evaluate on three tasks, missing-value imputation, future prediction and time interpolation, rather than on a partition, so none of their reported numbers is a clustering score. They also train under the GSNN objective \citep{Sohn2015}, which sets $q_\psi(z\mid Y)\equiv p(z\mid A,x)$ so the decoder only ever sees prior codes and the train/test gap closes by construction.

% =========================================================================
\section{Experimental Details}
\label{app:exp_det}
% =========================================================================

\subsection{Evaluation Metrics}
\label{app:metrics}

\paragraph{Clustering.} We report the adjusted Rand index (ARI) between the classes assigned to held-out subjects and their reference labels. ARI is $0$ in expectation for a random partition and $1$ for exact recovery, and it is invariant to label permutation. Every method is given the true number of classes, so no method is penalised for choosing $G$ badly.

\paragraph{Reconstruction.} We report the mean squared error (MSE) at hidden cells, on markers standardised with their observed training cells. Imputers return a completed panel and generative models a reconstruction; scoring hidden cells only puts both on the same footing. A mean fill then scores about $1$, so an entry reads as the share of a marker's variance left unexplained. LCMM-VAE reconstructs by \eqref{eq:hybrid}.

\paragraph{Downstream utility.} PhysioNet has no reference partition, so we follow \citet{Fortuin2020}: each completed stay, with its observed cells restored, is flattened to one $48\times35$ vector, a logistic regression fitted on half of the test stays predicts in-hospital mortality, and we report its AUROC on the other half. Chance is $0.5$ and the mean fill scores $0.692$. The score assesses the completion, not the partition, so methods without classes take part.

\paragraph{Paired tests.} Each method is compared with LCMM-VAE by a two-sided Wilcoxon signed-rank test over the seeds both scored, at which both saw the same data, mask and held-out subjects \citep{Demsar2006}. Holm's correction \citep{Holm1979} runs over all tests on one dataset (\tableref{tab:paired,tab:paired_rec}).

\subsection{Simulation}
\label{app:data_protocols}
\label{app:mech_all}
\label{app:mech}
\label{app:exp_sim}

\paragraph{Data.} We design a synthetic experiment to test whether a method recovers trajectory-based classes from many markers observed at sparse visits, when each subject deviates from its class. At the latent level, we generate a mixture of class-specific mixed models with prevalences $\pi=(0.5,0.3,0.2)$,
\begin{equation*}
  z_i(t)=\mu_{c_i}(t)+A_{R,i}\,x_R(t)+\varepsilon_i(t),
\end{equation*}
with $\varepsilon_i(t)\sim\Normal(\zerov{L},I_L)$, $z_i(t)\in\R^L$, $L=3$, and $t$ on a grid of $100$ points in $[0,1]$. We envision a physiological process in which subjects differ in \emph{when} it begins, \emph{what shape} it takes as it passes, and \emph{how fast} it then declines; for simplicity, these are the three latent coordinates. With $s(u)=1/(1+e^{-u})$ and $\tau_g=(0.35,0.50,0.65)$, the coordinates $\mu_{g,l}$, $l=1,2,3$, of the class curves $\mu_g$ are, up to a common affine rescaling,
\begin{equation*}
\begin{split}
  \mu_{g,1}(t)&=s\bigl(18(t-\tau_g)\bigr),\qquad
  \mu_{g,2}(t)=e^{-((t-\tau_g)/0.2)^2},\\
  \mu_{g,3}(t)&=1-0.8\,s\bigl(6(t-0.5)\bigr)-0.3\,(g-2)\,t.
\end{split}
\end{equation*}
Each subject draws a random intercept and slope on every coordinate, $x_R(t)=(1,t)^{\top}$ and $a_{R,i,l}\sim\Normal(\zerov{2},\diag(0.5,1.5))$. We decode $10$ informative markers as $Y^{\mathrm{sig}}_i(t)=f\bigl(z_i(t)\bigr)+0.5\,h\bigl(v_i(t)\bigr)$, where $f(z)=W_2\tanh(W_1z+b_1)+b_2$, $f:\R^3\to\R^{10}$, is a random network of hidden width $32$, fixed and never trained, and the smooth noise passes a $3$-dimensional Gaussian process $v_i$ (RBF kernel, length scale $0.15$) through a second such network $h$, both terms standardised. We add $40$ nuisance markers that carry no class information, each an independent draw of the same Gaussian process, and standardise every marker. Each subject keeps about $20\%$ of the grid: a grid point is visited at random (MCAR) or with a probability that increases with the last observed value of an informative marker (MAR), at most $19$ points are skipped in a row, and a missed visit hides every marker.

\paragraph{Protocol.} We draw $N=500$ subjects ($1000$ in \tableref{tab:sim_full}), hold out $15\%$, and score reconstruction at their dropped grid points on the $10$ informative markers. Every method is given $G=3$, and every latent-variable method $L=3$.

\subsection{Gait Data}
\label{app:exp_gait}
\label{app:gait_data}
\label{app:gait_splits}
\label{app:gait_protocol}

\paragraph{Data.} The Multivariate Gait Data \citep{Helwig2016} record $10$ healthy subjects walking with no brace, a knee brace and an ankle brace, $10$ consecutive gait cycles each, time-normalised to $101$ points, with the joint angles of the ankle, knee and hip on both legs ($K=6$). A walk is one cycle, labelled by its condition.

\paragraph{Split.} Each seed holds out two of the ten persons, with all $60$ of their walks, and trains on the walks of the other eight. All possible pairs are shuffled once, and seed $s$ holds out the $s$-th. Markers are standardised with training statistics.

\paragraph{Missingness.} The data are complete, so missingness is imposed by two mechanisms, each at three rates. \texttt{mcar} hides each cell independently with probability $0.2$, $0.4$ or $0.6$, so a hidden run is short. \texttt{channel} removes whole markers from a walk, at rate $p=0.17$, $0.33$ or $0.50$, always keeping at least one. No temporal neighbour then survives, every local rule falls back to the marker mean, and a missing marker can only be recovered across markers, through a shared latent.

\subsection{PhysioNet Data}
\label{app:exp_physionet_det}
\label{app:pn_data}
\label{app:impl}
\label{app:pn_splits}

\paragraph{Data.} We use the extraction released by \citet{Fortuin2020}: the first $48$\,h of each ICU stay, binned hourly into a $48\times35$ panel with $82\%$ of cells missing, in fixed training, validation and test splits.

\paragraph{Task.} Mortality prediction from the completed series (\appendixref{app:metrics}). Clustering-only methods produce no completion and have no row.

\subsection{Implementation}
\label{app:ourcfg}
\label{app:gait_ourcfg}
\label{app:gait_setup}
\label{app:hpsearch}

\paragraph{LCMM-VAE.} \tableref{tab:ourcfg} gives the configuration, where the encoder widths, pretraining length, the learning rate and $S$ were selected per dataset by a grid search on validation subjects (scored by the MSE at artificially hidden cells). $L$ is shared by every latent-variable method rather than searched: the generator's $L=3$ on the simulation, and $L=\min(10,K)$ on real data, so that $K\ge L$ as Lemma~\ref{lem:decoder} requires (\appendixref{app:latent_dim} varies it). On gait and PhysioNet, $R=0$ (\sectionref{sec:discussion}).

\begin{table}[H]
\centering
\caption{\textbf{LCMM-VAE configuration}. B-splines are cubic.}
\label{tab:ourcfg}
\footnotesize\setlength{\tabcolsep}{3pt}
\begin{tabular*}{\linewidth}{@{}l@{\extracolsep{\fill}}lll@{}}
\toprule
& Simulation & Gait & PhysioNet \\
\midrule
$K$ / $L$ / $G$ & $50$ / $3$ / $3$ & $6$ / $6$ / $3$ & $35$ / $10$ / $4$ \\
$x_S$ & $6$ B-splines & $12$ B-splines & $6$ B-splines \\
$x_R$ & $(1,t)$, $R=2$ & none & none \\
\midrule
encoder widths & $[256,128]$ & $[512,256]$ & $[128,64]$ \\
pretraining & $10$ epochs & $20$ epochs & $10$ epochs \\
learning rate & $10^{-3}$ & $10^{-3}$ & $3\times10^{-4}$ \\
\midrule
activation & \multicolumn{3}{>{\raggedright\arraybackslash}p{0.66\linewidth}}{ELU} \\
variances & \multicolumn{3}{>{\raggedright\arraybackslash}p{0.66\linewidth}}{$\sigma_\varepsilon^2=1$; $\sigma_y^2$ per marker, initial value $1$, floor $3\times10^{-4}$} \\
KL warm-up & \multicolumn{3}{>{\raggedright\arraybackslash}p{0.66\linewidth}}{linear} \\
initialisation & \multicolumn{3}{>{\raggedright\arraybackslash}p{0.66\linewidth}}{em-EM, $50$ runs of $\le5$ iterations, best run continued for $\le200$ iterations (Rmixmod's defaults)} \\
training & \multicolumn{3}{>{\raggedright\arraybackslash}p{0.66\linewidth}}{$\beta=1$; Adam, batch $512$; $\le50$ epochs, stop when $<0.1\%$ of subjects change class \citep{Xie2016}} \\
inference & \multicolumn{3}{>{\raggedright\arraybackslash}p{0.66\linewidth}}{$20$ E-step sweeps $m_{i,l}\to\gamma_i$ from $\gamma_i=\pi$} \\
\bottomrule
\end{tabular*}
\end{table}

\paragraph{Baselines.}
\label{app:baselines}
\label{app:gait_baselines}
\label{app:authorcode}
Each baseline runs its authors' code or a standard library (PyPOTS~1.5, scikit-learn, kshape-python, \texttt{lcmm}~2.2.2) at its published configuration for the dataset where one exists, and otherwise at its code's defaults. Latent-variable methods take the dataset's $L$ unless their authors published a configuration for it; SOM-VAE keeps $\dim(z)=64$, as its encoder degenerates at small $L$. Every $k$-means or mixture step we add keeps the best of $50$ restarts, and GP-VAE and SOM-VAE label subjects by $k$-means on their latent means. Methods that need a complete grid receive the training-mean fill. Training is capped at $2.5$\,h per fit (CRLI at $200$ epochs, SAITS at $300$, BRITS at $700$), and the LCMM, run on gait only, at one start and $6$\,h. For the two-stage spline methods, $S$ and the ridge of the per-subject fits are searched like LCMM-VAE's settings, scored by the error of those fits; the selected setting, used for their PhysioNet reconstruction, lowers their clustering, so we report their clustering at the basis used before the search ($S=6$ on the simulation and $10$ on gait), the more favourable to them (\appendixref{app:basis}).

\paragraph{Seeds and compute.}
\label{app:protocol}
We run $30$ seeds per cell, or $10$ for a method whose first fit on a dataset exceeds $500$\,s. A seed fixes the simulated cohort or held-out pair, the mask, the initialisation and the data order; on PhysioNet, whose split is fixed, only the last two. A failed fit is reported, never retried: VaDE failed $2$ of $30$ fits on the simulation (MAR, $N=1000$) and VaDER $4$ of $30$ on PhysioNet. All fits ran on one CPU host (Intel Core i9-9820X); SAITS ran on its GPU (NVIDIA RTX~3090), so its times, and BRITS's on PhysioNet, are not reported (n.r.). Times are median wall-clock seconds, including contention between concurrent fits, and indicate scale rather than exact cost.

% =========================================================================

\section{Additional Results}
\label{app:extra}
% =========================================================================

\FloatBarrier
\subsection{Statistical Tests}
\label{app:paired}
\tableref{tab:paired,tab:paired_rec} compare every baseline and LCMM-VAE ($G{=}1$) with LCMM-VAE by the paired tests of \appendixref{app:metrics}, on the columns of \tableref{tab:main,tab:recon} and on the PhysioNet AUROC (\appendixref{app:pn_auroc}).

\subsection{Full Results}
\label{app:full_results}
\label{app:gait_extra}
This section reports every method at every setting: the simulation clustering at both cohort sizes, and gait at every (mechanism, rate).

\paragraph{Simulation.} \tableref{tab:sim_full} adds $N=1000$ and four methods absent from \tableref{tab:main}: $k$-means on the flattened panel, $k$-Shape \citep{Paparrizos2015}, DTCR and SOM-VAE $+$ $k$-means. Doubling $N$ moves LCMM-VAE from $0.90$ to $0.93$ (MCAR) and from $0.76$ to $0.81$ (MAR). Its $G{=}1$ ablation follows ($0.90$, $0.73$) and stays below LCMM-VAE in every cell; at $N=500$ the gap is significant under MAR only (\tableref{tab:paired}). Every deep clustering baseline stays below $0.25$ at both sizes.

\paragraph{Gait.} \tableref{tab:gait_grid_full} gives ARI at every (mechanism, rate) and \tableref{tab:gait_recon_rates} the reconstruction. Under \texttt{mcar}, LCMM-VAE's ARI decreases from $0.89$ at $20\%$ to $0.75$ at $60\%$, while the two-stage methods stay within $0.87$--$0.97$. Under \texttt{channel}, it has the highest ARI at $p=0.33$ ($0.59$) and the lowest reconstruction error at $p=0.17$ and $0.33$ ($0.268$ and $0.322$); at $p=0.50$ its ARI falls to $0.15$, and its $G{=}1$ ablation reconstructs better ($0.357$ against $0.467$). \figureref{fig:recon} shows one held-out walk under each mechanism.

\subsection{Sensitivity Analyses}
\label{app:sensitivity}
\paragraph{Latent dimension.}
\label{app:latent_dim}
$L$ is shared by every latent-variable method rather than searched (\appendixref{app:hpsearch}). \tableref{tab:sim_L10} refits every method that shares $L$ at $L=10$ on the simulation, seeds $0$--$9$. LCMM-VAE keeps its ARI under MCAR ($0.89$ at both $L$) but loses under MAR ($0.57$ against $0.77$), while staying above its $G{=}1$ ablation and every other method.

\paragraph{Basis size.}
\label{app:basis}
\figureref{fig:basis} varies the number of B-spline functions $S$ on the simulation for LCMM-VAE and every method that uses the basis, all other settings as in the main-text run ($S=6$). LCMM-VAE needs enough basis functions to represent the class curves: at $S=4$, the smallest cubic basis (a single cubic polynomial), its ARI falls to $0.80$ under MCAR and $0.56$ under MAR. From $S=6$ on its ARI is stable, moving by at most $0.07$. The B-spline methods peak at $S=6$ and collapse as $S$ grows, because each subject's roughly $20$ visits no longer determine its own coefficients, whereas LCMM-VAE estimates the class coefficients from all the subjects of a class. The validation search (\appendixref{app:hpsearch}) selects a larger basis for them, jointly with the ridge of their spline fits ($S=10$, ridge $0.1$), at which their clustering degrades further; we therefore report them at $S=6$, the setting most favourable to them.

\subsection{Selecting the Number of Classes}
\label{app:select_g}
In practice, $G$ would be chosen as for latent-class mixed models \citep{ProustLima2017}, by comparing BIC or ICL \citep{Biernacki2000} across fits with different $G$. We test this on the simulation without labels: on each of the $60$ cohorts of \tableref{tab:main}, we freeze the pretrained encoder and fit the latent LCMM of the initialisation (\sectionref{sec:init}) for $G=1,\dots,5$ on the training encoder means. BIC recovers $G=3$ in $29$ of $30$ cohorts under MCAR and $20$ under MAR, where the classes overlap more; ICL, which penalises uncertain assignments, more often merges two classes (\tableref{tab:select_g}). As for any mixture model, candidate partitions are best compared rather than chosen by a single criterion.

\subsection{PhysioNet Mortality Prediction}
\label{app:pn_auroc}
\tableref{tab:physionet} reports the mortality-prediction benchmark of \citet{Fortuin2020} (\appendixref{app:metrics}). LCMM-VAE ($0.703$) is significantly above the mean fill ($0.692$) and CRLI ($0.636$), comparable to VaDER ($0.709$), and significantly below the other baselines, up to $0.728$ for SAITS. The extraction of \citet{Fortuin2020} has two costs for our model: it is already binned, so LCMM-VAE never sees the irregular timestamps it is built for, and its only covariate is time, where \citet{Ong2025} also supply age, gender and ICU type, so our classes are functions of time alone.

\subsection{PhysioNet Classes}
\label{app:classes}
\label{app:pn_extra}
The classes are those of the reference seed, the seed whose partition has the highest mean ARI to the other nine, fitted on the training stays and described on the test stays. \tableref{tab:pn_classes} describes them on variables the model never sees.

\clearpage
% ---- Appendix H tables and figures, after its text
\begin{table*}[tp]
\centering
\caption{\textbf{Paired Wilcoxon signed-rank tests, clustering} (ARI, simulation at $N=500$). Two-sided Holm-corrected tests over the seeds each pair shares; cell: median paired difference, oriented so that a positive value favours LCMM-VAE. $^{*}p<0.05$, $^{**}p<0.01$, $^{***}p<0.001$; n.s.: not significant after correction. Gait columns test per-seed means over the three rates.}
\label{tab:paired}
\small\setlength{\tabcolsep}{5pt}
\begin{tabular}{@{}l cc ccc@{}}
\toprule
 & \multicolumn{2}{c}{Simulation} & \multicolumn{3}{c}{Gait} \\
\cmidrule(lr){2-3}\cmidrule(lr){4-6}
Method & MCAR & MAR & complete & \texttt{mcar} & \texttt{channel} \\
\midrule
$k$-means (B-spline) & $+0.21^{***}$ & $+0.19^{***}$ & $+0.00$ n.s. & $-0.08$ n.s. & $+0.29^{***}$ \\
GMM-diag (B-spline) & $+0.17^{***}$ & $+0.33^{***}$ & $+0.00$ n.s. & $-0.11^{**}$ & $+0.42^{***}$ \\
FPCA $+$ $k$-means & $+0.64^{***}$ & $+0.56^{***}$ & $+0.00$ n.s. & $-0.05^{*}$ & $+0.06^{**}$ \\
\midrule
$k$-means (flattened panel) & $+0.46^{***}$ & $+0.77^{***}$ & $+0.00$ n.s. & $-0.09^{**}$ & $+0.05^{**}$ \\
$k$-Shape & $+0.87^{*}$ & $+0.76^{*}$ & $+0.44^{***}$ & $+0.32^{***}$ & $+0.04$ n.s. \\
IDEC & $+0.79^{***}$ & $+0.81^{***}$ & $+0.06^{*}$ & $+0.12^{**}$ & $0.00$ n.s. \\
VaDE & $+0.91^{***}$ & $+0.81^{***}$ & $+0.35^{**}$ & $+0.23^{***}$ & $+0.14^{***}$ \\
DTCR & $+0.89^{*}$ & $+0.78^{*}$ & $+0.52^{*}$ & $+0.52^{*}$ & $+0.37^{*}$ \\
SOM-VAE $+$ $k$-means & $+0.71^{*}$ & $+0.71^{*}$ & $+0.04$ n.s. & $+0.00$ n.s. & $+0.31^{*}$ \\
\midrule
VaDER & $+0.89^{***}$ & $+0.81^{***}$ & $+0.65^{***}$ & $+0.73^{***}$ & $+0.46^{***}$ \\
CRLI & $+0.87^{*}$ & $+0.79^{*}$ & $+0.19$ n.s. & $+0.05$ n.s. & $+0.04$ n.s. \\
\midrule
GP-VAE $+$ $k$-means & $+0.65^{***}$ & $+0.70^{***}$ & $+0.06^{**}$ & $+0.29^{***}$ & $+0.09^{***}$ \\
LCMM-VAE ($G{=}1$, ablation) & $+0.01$ n.s. & $+0.06^{*}$ & $+0.00$ n.s. & $+0.19^{***}$ & $+0.10^{***}$ \\
\bottomrule
\end{tabular}
\end{table*}

\begin{table*}[tp]
\centering
\caption{\textbf{Paired Wilcoxon signed-rank tests, reconstruction} (MSE; PhysioNet: mortality AUROC), as in \tableref{tab:paired}.}
\label{tab:paired_rec}
\label{tab:paired_pn}
\small\setlength{\tabcolsep}{5pt}
\begin{tabular}{@{}l cc cc c@{}}
\toprule
 & \multicolumn{2}{c}{Simulation} & \multicolumn{2}{c}{Gait} & PhysioNet \\
\cmidrule(lr){2-3}\cmidrule(lr){4-5}\cmidrule(lr){6-6}
Method & MCAR & MAR & \texttt{mcar} & \texttt{channel} & AUROC \\
\midrule
Mean fill & $+0.409^{***}$ & $+0.526^{***}$ & $+0.753^{***}$ & $+0.687^{***}$ & $+0.011^{*}$ \\
LOCF & $+0.143^{***}$ & $+0.197^{***}$ & $-0.230^{***}$ & $+0.687^{***}$ & $-0.017^{*}$ \\
Linear interp. & $-0.040^{***}$ & $-0.058^{***}$ & $-0.273^{***}$ & $+0.687^{***}$ & $-0.015^{*}$ \\
B-spline fit (two-stage) & -- & -- & -- & -- & $-0.009^{*}$ \\
\midrule
VaDER & -- & -- & -- & -- & $+0.003$ n.s. \\
CRLI & -- & -- & -- & -- & $+0.068^{*}$ \\
\midrule
SAITS & $+0.178^{*}$ & $+0.438^{*}$ & $-0.260^{*}$ & $+0.321^{*}$ & $-0.026^{*}$ \\
BRITS & $+0.009$ n.s. & $+0.099^{*}$ & $-0.267^{*}$ & $+0.239^{*}$ & $-0.018^{*}$ \\
GP-VAE & $+0.156^{***}$ & $+0.151^{***}$ & $-0.065^{***}$ & $+0.121^{***}$ & $-0.007^{*}$ \\
LCMM-VAE ($G{=}1$, ablation) & $+0.090^{***}$ & $+0.131^{***}$ & $+0.095^{***}$ & $-0.012$ n.s. & $+0.001^{*}$ \\
\bottomrule
\end{tabular}
\end{table*}

\begin{table*}[tp]
\centering
\caption{\textbf{Simulation: clustering}, held-out ARI, $\text{mean}_{\pm\text{s.d.}}$ over seeds. Bold: best mean per column.}
\label{tab:sim_full}
\small\setlength{\tabcolsep}{5pt}
\begin{tabular}{@{}l cc cc@{}}
\toprule
& \multicolumn{2}{c}{MCAR} & \multicolumn{2}{c}{MAR} \\
\cmidrule(lr){2-3}\cmidrule(lr){4-5}
Method & $N=500$ & $N=1000$ & $N=500$ & $N=1000$ \\
\midrule
$k$-means (B-spline) & $0.68_{\pm 0.16}$ & $0.71_{\pm 0.12}$ & $0.58_{\pm 0.18}$ & $0.62_{\pm 0.16}$ \\
GMM-diag (B-spline) & $0.71_{\pm 0.14}$ & $0.74_{\pm 0.13}$ & $0.44_{\pm 0.14}$ & $0.48_{\pm 0.09}$ \\
FPCA $+$ $k$-means & $0.28_{\pm 0.07}$ & $0.28_{\pm 0.05}$ & $0.28_{\pm 0.06}$ & $0.27_{\pm 0.05}$ \\
\midrule
$k$-means (flattened panel) & $0.41_{\pm 0.14}$ & $0.36_{\pm 0.13}$ & $0.04_{\pm 0.09}$ & $0.09_{\pm 0.12}$ \\
$k$-Shape & $0.01_{\pm 0.02}$ & $0.00_{\pm 0.01}$ & $0.02_{\pm 0.03}$ & $0.01_{\pm 0.03}$ \\
IDEC & $0.13_{\pm 0.13}$ & $0.22_{\pm 0.12}$ & $0.01_{\pm 0.04}$ & $0.08_{\pm 0.10}$ \\
VaDE & $0.00_{\pm 0.00}$ & $0.02_{\pm 0.06}$ & $0.01_{\pm 0.05}$ & $0.08_{\pm 0.09}$ \\
DTCR & $0.00_{\pm 0.02}$ & $0.01_{\pm 0.03}$ & $0.00_{\pm 0.03}$ & $0.00_{\pm 0.03}$ \\
SOM-VAE $+$ $k$-means & $0.22_{\pm 0.11}$ & $0.24_{\pm 0.06}$ & $0.04_{\pm 0.08}$ & $0.08_{\pm 0.09}$ \\
\midrule
VaDER & $0.03_{\pm 0.06}$ & $0.09_{\pm 0.10}$ & $0.03_{\pm 0.07}$ & $0.07_{\pm 0.10}$ \\
CRLI & $0.02_{\pm 0.03}$ & $0.01_{\pm 0.04}$ & $0.02_{\pm 0.05}$ & $0.00_{\pm 0.03}$ \\
\midrule
GP-VAE $+$ $k$-means & $0.24_{\pm 0.10}$ & $0.24_{\pm 0.07}$ & $0.08_{\pm 0.08}$ & $0.15_{\pm 0.10}$ \\
LCMM-VAE ($G{=}1$, ablation) & $0.88_{\pm 0.08}$ & $0.90_{\pm 0.10}$ & $0.69_{\pm 0.19}$ & $0.73_{\pm 0.19}$ \\
\midrule
LCMM-VAE (ours, $R{=}0$) & $0.29_{\pm 0.10}$ & $0.29_{\pm 0.04}$ & $0.27_{\pm 0.08}$ & $0.29_{\pm 0.10}$ \\
LCMM-VAE (ours, $R{=}2$) & $\mathbf{0.90}_{\pm 0.08}$ & $\mathbf{0.93}_{\pm 0.07}$ & $\mathbf{0.76}_{\pm 0.20}$ & $\mathbf{0.81}_{\pm 0.19}$ \\
\bottomrule
\end{tabular}
\end{table*}

\begin{table*}[tp]
\centering
\caption{\textbf{Gait: clustering per missingness rate}, held-out ARI, $\text{mean}_{\pm\text{s.d.}}$ over seeds; \texttt{channel} columns give $p$. Bold: best mean per column. \emph{in}: \native{} consumes the mask, \imputed{} needs a complete tensor.}
\label{tab:gait_grid_full}
\small\setlength{\tabcolsep}{4pt}
\begin{tabular}{@{}lc c ccc ccc@{}}
\toprule
& & & \multicolumn{3}{c}{\texttt{mcar}} & \multicolumn{3}{c}{\texttt{channel}} \\
\cmidrule(lr){4-6}\cmidrule(lr){7-9}
Method & in & complete & $20\%$ & $40\%$ & $60\%$ & $0.17$ & $0.33$ & $0.50$ \\
\midrule
$k$-means (B-spline) & \native & $0.90_{\pm 0.14}$ & $0.90_{\pm 0.14}$ & $0.88_{\pm 0.16}$ & $0.91_{\pm 0.11}$ & $0.50_{\pm 0.20}$ & $0.17_{\pm 0.07}$ & $0.03_{\pm 0.06}$ \\
GMM-diag (B-spline) & \native & $0.97_{\pm 0.10}$ & $\mathbf{0.97}_{\pm 0.09}$ & $\mathbf{0.96}_{\pm 0.11}$ & $0.91_{\pm 0.11}$ & $0.33_{\pm 0.12}$ & $0.00_{\pm 0.02}$ & $0.01_{\pm 0.03}$ \\
FPCA $+$ $k$-means & \native & $0.94_{\pm 0.12}$ & $0.93_{\pm 0.12}$ & $0.93_{\pm 0.12}$ & $0.87_{\pm 0.18}$ & $0.75_{\pm 0.11}$ & $0.44_{\pm 0.13}$ & $0.17_{\pm 0.04}$ \\
\midrule
$k$-means (flattened panel) & \imputed & $0.94_{\pm 0.12}$ & $0.94_{\pm 0.11}$ & $0.94_{\pm 0.10}$ & $\mathbf{0.93}_{\pm 0.10}$ & $0.76_{\pm 0.09}$ & $0.44_{\pm 0.13}$ & $0.17_{\pm 0.04}$ \\
$k$-Shape & \imputed & $0.55_{\pm 0.09}$ & $0.55_{\pm 0.06}$ & $0.53_{\pm 0.05}$ & $0.46_{\pm 0.05}$ & $0.54_{\pm 0.10}$ & $0.48_{\pm 0.06}$ & $\mathbf{0.40}_{\pm 0.08}$ \\
IDEC & \imputed & $0.83_{\pm 0.18}$ & $0.79_{\pm 0.18}$ & $0.74_{\pm 0.18}$ & $0.67_{\pm 0.18}$ & $0.76_{\pm 0.17}$ & $0.51_{\pm 0.13}$ & $0.28_{\pm 0.10}$ \\
VaDE & \imputed & $0.68_{\pm 0.26}$ & $0.59_{\pm 0.29}$ & $0.56_{\pm 0.23}$ & $0.60_{\pm 0.16}$ & $0.47_{\pm 0.21}$ & $0.31_{\pm 0.22}$ & $0.22_{\pm 0.10}$ \\
DTCR & \imputed & $0.38_{\pm 0.19}$ & $0.33_{\pm 0.18}$ & $0.32_{\pm 0.17}$ & $0.29_{\pm 0.17}$ & $0.25_{\pm 0.12}$ & $0.09_{\pm 0.09}$ & $0.06_{\pm 0.09}$ \\
SOM-VAE $+$ $k$-means & \imputed & $0.85_{\pm 0.18}$ & $0.85_{\pm 0.16}$ & $0.84_{\pm 0.16}$ & $0.86_{\pm 0.13}$ & $0.44_{\pm 0.14}$ & $0.14_{\pm 0.10}$ & $0.03_{\pm 0.05}$ \\
\midrule
VaDER & \native & $0.23_{\pm 0.22}$ & $0.19_{\pm 0.20}$ & $0.13_{\pm 0.17}$ & $0.15_{\pm 0.18}$ & $0.11_{\pm 0.15}$ & $0.10_{\pm 0.14}$ & $0.08_{\pm 0.13}$ \\
CRLI & \native & $0.73_{\pm 0.30}$ & $0.75_{\pm 0.30}$ & $0.75_{\pm 0.36}$ & $0.75_{\pm 0.28}$ & $0.56_{\pm 0.28}$ & $0.43_{\pm 0.15}$ & $0.39_{\pm 0.15}$ \\
\midrule
GP-VAE $+$ $k$-means & \native & $0.81_{\pm 0.20}$ & $0.71_{\pm 0.25}$ & $0.47_{\pm 0.23}$ & $0.37_{\pm 0.15}$ & $0.64_{\pm 0.21}$ & $0.35_{\pm 0.18}$ & $0.19_{\pm 0.09}$ \\
LCMM-VAE ($G{=}1$, ablation) & \native & $\mathbf{1.00}_{\pm 0.00}$ & $0.96_{\pm 0.06}$ & $0.62_{\pm 0.22}$ & $0.38_{\pm 0.12}$ & $\mathbf{0.83}_{\pm 0.16}$ & $0.39_{\pm 0.12}$ & $0.04_{\pm 0.05}$ \\
\midrule
LCMM-VAE (ours, $R{=}0$) & \native & $0.94_{\pm 0.10}$ & $0.89_{\pm 0.16}$ & $0.87_{\pm 0.15}$ & $0.75_{\pm 0.18}$ & $0.79_{\pm 0.13}$ & $\mathbf{0.59}_{\pm 0.17}$ & $0.15_{\pm 0.10}$ \\
\bottomrule
\end{tabular}
\end{table*}

\begin{table*}[tp]
\centering
\caption{\textbf{Gait: reconstruction per missingness rate}, MSE at hidden cells, $\text{mean}_{\pm\text{s.d.}}$ over seeds. Bold: best mean per column.}
\label{tab:gait_recon_rates}
\small\setlength{\tabcolsep}{5pt}
\begin{tabular}{@{}l ccc ccc@{}}
\toprule
& \multicolumn{3}{c}{\texttt{mcar}} & \multicolumn{3}{c}{\texttt{channel}} \\
\cmidrule(lr){2-4}\cmidrule(lr){5-7}
Method & $20\%$ & $40\%$ & $60\%$ & $0.17$ & $0.33$ & $0.50$ \\
\midrule
Mean fill & $1.024_{\pm 0.066}$ & $1.024_{\pm 0.067}$ & $1.027_{\pm 0.066}$ & $1.019_{\pm 0.080}$ & $1.032_{\pm 0.082}$ & $1.030_{\pm 0.075}$ \\
LOCF & $0.019_{\pm 0.003}$ & $0.038_{\pm 0.005}$ & $0.091_{\pm 0.014}$ & $1.019_{\pm 0.080}$ & $1.032_{\pm 0.082}$ & $1.030_{\pm 0.075}$ \\
Linear interp. & $\mathbf{0.001}_{\pm 0.000}$ & $\mathbf{0.002}_{\pm 0.000}$ & $\mathbf{0.009}_{\pm 0.002}$ & $1.019_{\pm 0.080}$ & $1.032_{\pm 0.082}$ & $1.030_{\pm 0.075}$ \\
\midrule
SAITS & $0.007_{\pm 0.002}$ & $0.015_{\pm 0.003}$ & $0.030_{\pm 0.007}$ & $0.623_{\pm 0.089}$ & $0.622_{\pm 0.113}$ & $0.676_{\pm 0.084}$ \\
BRITS & $0.004_{\pm 0.001}$ & $0.005_{\pm 0.001}$ & $0.011_{\pm 0.002}$ & $0.634_{\pm 0.141}$ & $0.545_{\pm 0.087}$ & $0.626_{\pm 0.179}$ \\
GP-VAE & $0.141_{\pm 0.033}$ & $0.214_{\pm 0.048}$ & $0.274_{\pm 0.036}$ & $0.397_{\pm 0.060}$ & $0.475_{\pm 0.061}$ & $0.536_{\pm 0.052}$ \\
LCMM-VAE ($G{=}1$, ablation) & $0.335_{\pm 0.037}$ & $0.355_{\pm 0.031}$ & $0.422_{\pm 0.031}$ & $0.317_{\pm 0.050}$ & $0.339_{\pm 0.061}$ & $\mathbf{0.357}_{\pm 0.034}$ \\
\midrule
LCMM-VAE (ours, $R{=}0$) & $0.214_{\pm 0.028}$ & $0.258_{\pm 0.026}$ & $0.359_{\pm 0.034}$ & $\mathbf{0.268}_{\pm 0.056}$ & $\mathbf{0.322}_{\pm 0.054}$ & $0.467_{\pm 0.064}$ \\
\bottomrule
\end{tabular}
\end{table*}

\begin{figure*}[tp]
\floatconts
  {fig:recon}
  {\caption{\textbf{Gait: reconstruction of one held-out walk} (left knee, seed $0$, drawn at random), under \texttt{mcar} $40\%$ (\emph{left}) and \texttt{channel} $p=0.33$ (\emph{right}). Dots: observed cells; circles: hidden cells. Values in each panel: MSE at this walk's hidden cells.}}
  {\centering\includegraphics[width=\linewidth]{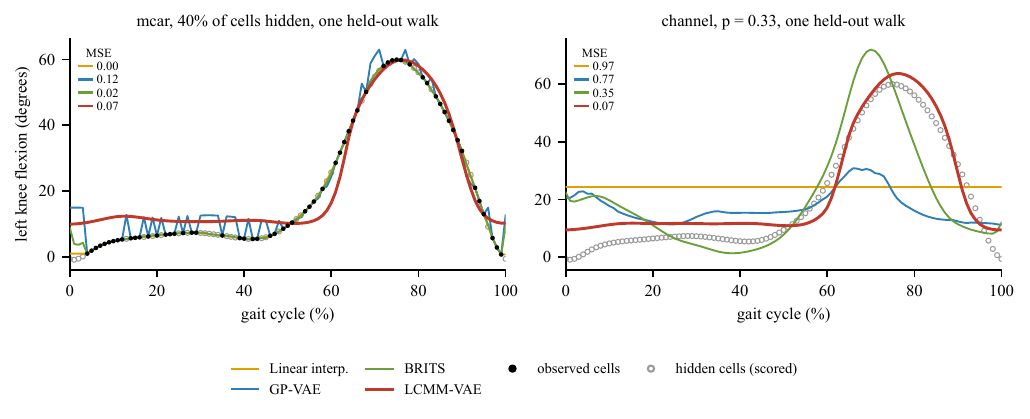}}
\end{figure*}

\begin{table*}[tp]
\centering
\caption{\textbf{Sensitivity analysis: latent dimension (simulation)}, held-out ARI, $N=500$, $\text{mean}_{\pm\text{s.d.}}$ over seeds $0$--$9$.}
\label{tab:sim_L10}
\small\setlength{\tabcolsep}{5pt}
\begin{tabular}{@{}l cc cc@{}}
\toprule
& \multicolumn{2}{c}{MCAR} & \multicolumn{2}{c}{MAR} \\
\cmidrule(lr){2-3}\cmidrule(lr){4-5}
Method & $L=3$ & $L=10$ & $L=3$ & $L=10$ \\
\midrule
IDEC & $0.17_{\pm 0.16}$ & $0.21_{\pm 0.17}$ & $0.01_{\pm 0.03}$ & $0.11_{\pm 0.13}$ \\
VaDE & $0.00_{\pm 0.00}$ & $0.00_{\pm 0.00}$ & $0.01_{\pm 0.02}$ & $0.00_{\pm 0.00}$ \\
VaDER & $0.02_{\pm 0.05}$ & $0.00_{\pm 0.03}$ & $0.01_{\pm 0.04}$ & $0.03_{\pm 0.05}$ \\
\midrule
GP-VAE $+$ $k$-means & $0.25_{\pm 0.13}$ & $0.23_{\pm 0.12}$ & $0.11_{\pm 0.08}$ & $0.11_{\pm 0.08}$ \\
LCMM-VAE ($G{=}1$, ablation) & $0.85_{\pm 0.08}$ & $0.81_{\pm 0.16}$ & $0.68_{\pm 0.10}$ & $0.50_{\pm 0.13}$ \\
\midrule
LCMM-VAE (ours, $R{=}2$) & $0.89_{\pm 0.06}$ & $0.89_{\pm 0.06}$ & $0.77_{\pm 0.15}$ & $0.57_{\pm 0.21}$ \\
\bottomrule
\end{tabular}
\end{table*}

\begin{table*}[tp]
\centering
\caption{\textbf{$G$ selection (simulation)}, $N=500$: cohorts, out of $30$ per mechanism, in which each criterion selects each $G$; the true value is $G=3$.}
\label{tab:select_g}
\small\setlength{\tabcolsep}{5pt}
\begin{tabular}{@{}ll ccccc@{}}
\toprule
& & $G=1$ & $2$ & $\mathbf{3}$ & $4$ & $5$ \\
\midrule
MCAR & BIC & $0$ & $1$ & $\mathbf{29}$ & $0$ & $0$ \\
     & ICL & $0$ & $6$ & $\mathbf{24}$ & $0$ & $0$ \\
MAR  & BIC & $0$ & $6$ & $\mathbf{20}$ & $4$ & $0$ \\
     & ICL & $0$ & $18$ & $\mathbf{12}$ & $0$ & $0$ \\
\bottomrule
\end{tabular}
\end{table*}

\begin{figure*}[tp]
\floatconts
  {fig:basis}
  {\caption{\textbf{Sensitivity analysis: basis size (simulation)}, held-out ARI, $N=500$, $\text{mean}_{\pm\text{s.e.}}$ over $30$ seeds.}}
  {\centering\includegraphics[width=\linewidth]{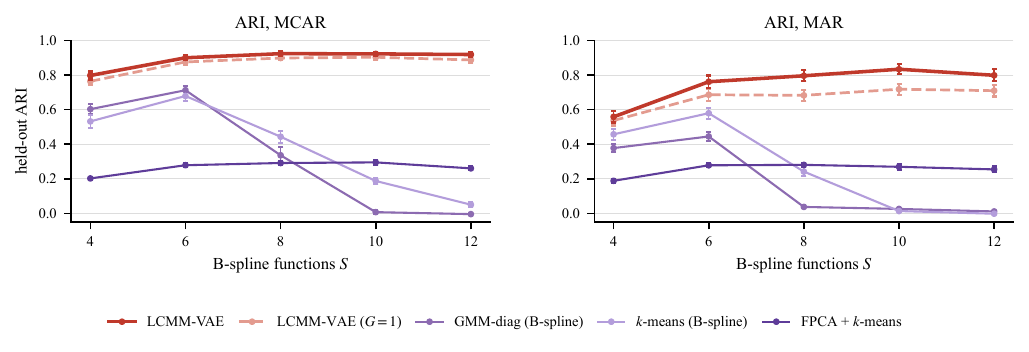}}
\end{figure*}

\begin{table*}[tp]
\centering
\caption{\textbf{PhysioNet: mortality AUROC} of a logistic regression on each method's completed test series \citep{Fortuin2020}, $\text{mean}_{\pm\text{s.d.}}$ over seeds. Bold: best mean per column. \emph{s}: median seconds per fit; n.r.: run on GPU, not reported.}
\label{tab:physionet}
\small\setlength{\tabcolsep}{5pt}
\begin{tabular}{@{}l c r@{}}
\toprule
Method & AUROC $\uparrow$ & s \\
\midrule
Mean fill & $0.692_{\pm 0.000}$ & 0.1 \\
LOCF & $0.720_{\pm 0.000}$ & 2.0 \\
Linear interp. & $0.717_{\pm 0.000}$ & 1.4 \\
B-spline fit (two-stage) & $0.712_{\pm 0.000}$ & 1.2 \\
\midrule
VaDER & $0.709_{\pm 0.009}$ & 286 \\
CRLI & $0.636_{\pm 0.023}$ & 8500 \\
\midrule
SAITS & $\mathbf{0.728}_{\pm 0.007}$ & n.r. \\
BRITS & $0.721_{\pm 0.007}$ & n.r. \\
GP-VAE & $0.710_{\pm 0.005}$ & 694 \\
LCMM-VAE ($G{=}1$, ablation) & $0.701_{\pm 0.001}$ & 875 \\
\midrule
LCMM-VAE (ours, $R{=}0$) & $0.703_{\pm 0.001}$ & 897 \\
\bottomrule
\end{tabular}
\end{table*}

\begin{table*}[tp]
\centering
\caption{\textbf{PhysioNet: LCMM-VAE classes on variables unseen in training}, test stays, reference seed; $\text{mean}_{\pm\text{s.d.}}$ or \%; $p$: ANOVA or $\chi^2$; on each of the ten seeds, age, ICU type, SAPS-I, SOFA, length of stay and in-hospital death have $p\le0.002$.}
\label{tab:pn_classes}\label{tab:pn_classes_short}
\small\setlength{\tabcolsep}{5pt}
\begin{tabular}{@{}l cccc c c@{}}
\toprule
 & A & B & C & D & all & $p$ \\
\midrule
stays & 1201 & 1195 & 948 & 653 & 3997 & \\
age (years) & $68.1_{\pm 15.7}$ & $69.2_{\pm 14.4}$ & $59.9_{\pm 17.6}$ & $57.3_{\pm 17.6}$ & $64.7_{\pm 16.8}$ & $<$0.001 \\
male (\%) & 56.3 & 55.5 & 54.0 & 57.6 & 55.7 & 0.52 \\
ICU: coronary care (\%) & 21 & 14 & 13 & 9 & 15 & $<$0.001 \\
ICU: cardiac-surgery recovery (\%) & 13 & 44 & 17 & 4 & 22 & \\
ICU: medical (\%) & 30 & 27 & 44 & 40 & 34 & \\
ICU: surgical (\%) & 35 & 15 & 26 & 47 & 29 & \\
SAPS-I & $14.0_{\pm 5.3}$ & $16.2_{\pm 5.0}$ & $15.9_{\pm 5.3}$ & $13.8_{\pm 4.8}$ & $15.1_{\pm 5.2}$ & $<$0.001 \\
SOFA & $5.3_{\pm 3.7}$ & $8.1_{\pm 3.7}$ & $7.8_{\pm 4.1}$ & $5.1_{\pm 3.6}$ & $6.7_{\pm 4.0}$ & $<$0.001 \\
length of stay (days) & $12.1_{\pm 11.2}$ & $12.5_{\pm 9.5}$ & $15.4_{\pm 16.3}$ & $13.9_{\pm 15.5}$ & $13.3_{\pm 13.0}$ & $<$0.001 \\
in-hospital death (\%) & 12.4 & 14.3 & 18.7 & 13.5 & 14.6 & $<$0.001 \\
\bottomrule
\end{tabular}
\end{table*}

\twocolumn[{%
\section{Related Methods}
\label{app:compare}
\tableref{tab:compare} tabulates, for related methods, which of the five properties of \sectionref{sec:rw} each has: a partition of the subjects, irregular visit times, many markers, covariates with individual deviations, and a trajectory per class that can be evaluated at any time. Rows are grouped by what each method returns. Univariate functional and mixed-model clustering \citep{JamesSugar2003, Shi2008, Nagin2005, Muthen2004} is not listed: like LCMM, it does not scale to many markers.

\medskip
\begin{minipage}{\textwidth}
\settabletype
\centering
\caption{\textbf{Related methods}, grouped by what they return, as published. \cmark~yes, \pmark~partly, \xmark~no, \dash~not applicable.}
\label{tab:compare}
\medskip
{\footnotesize\setlength{\tabcolsep}{2pt}
  \begin{tabular*}{\linewidth}{@{\extracolsep{\fill}}l c cc cc cc cc@{}}
  \toprule
   & & \multicolumn{2}{c}{Accepts} & \multicolumn{2}{c}{Models}
   & \multicolumn{2}{c}{Properties} & \multicolumn{2}{c}{Yields} \\
  \cmidrule(lr){3-4}\cmidrule(lr){5-6}\cmidrule(lr){7-8}\cmidrule(lr){9-10}
  Method & \rot{Partition} & \rot{Irregular times} & \rot{High-dim.\ $Y$}
   & \rot{Covariates} & \rot{Random effects} & \rot{Generative}
   & \rot{Identifiable} & \rot{Imputation} & \rot{Any $t$} \\
  \midrule
  \multicolumn{10}{@{}l}{\itshape Two-stage functional: per-subject splines, then clustering (class means only)}\\
  $k$-means (B-spline) \citep{Abraham2003}
    & \cmark & \cmark & \pmark & \xmark & \xmark & \xmark & \dash & \cmark & \cmark \\
  GMM-diag (B-spline)
    & \cmark & \cmark & \cmark & \xmark & \xmark & \pmark & \dash & \cmark & \cmark \\
  FPCA $+$ $k$-means \citep{Ramsay2005}
    & \cmark & \cmark & \cmark & \xmark & \xmark & \xmark & \dash & \cmark & \cmark \\
  \addlinespace
  \multicolumn{10}{@{}l}{\itshape Partition only}\\
  $k$-means (flattened panel) & \cmark & \xmark & \cmark & \xmark & \xmark & \xmark & \dash & \xmark & \xmark \\
  $k$-Shape \citep{Paparrizos2015} & \cmark & \xmark & \pmark & \xmark & \xmark & \xmark & \dash & \xmark & \xmark \\
  IDEC \citep{Guo2017} & \cmark & \xmark & \cmark & \xmark & \xmark & \xmark & \dash & \xmark & \xmark \\
  VaDE \citep{Jiang2017} & \cmark & \xmark & \cmark & \xmark & \xmark & \cmark & \xmark & \xmark & \xmark \\
  T-LSTM \citep{Baytas2017} & \cmark & \cmark & \cmark & \xmark & \xmark & \xmark & \dash & \xmark & \xmark \\
  DTCR \citep{MaDTCR2019} & \cmark & \xmark & \pmark & \xmark & \xmark & \xmark & \dash & \xmark & \xmark \\
  SOM-VAE \citep{Fortuin2019} & \pmark & \xmark & \cmark & \xmark & \xmark & \xmark & \dash & \xmark & \xmark \\
  T-DPSOM \citep{Manduchi2021} & \pmark & \xmark & \cmark & \xmark & \xmark & \cmark & \xmark & \xmark & \xmark \\
  \addlinespace
  \multicolumn{10}{@{}l}{\itshape Partition and reconstruction, but no class trajectory}\\
  VaDER \citep{deJong2019} & \cmark & \pmark & \cmark & \xmark & \xmark & \cmark & \xmark & \cmark & \xmark \\
  CRLI \citep{MaCRLI2021} & \cmark & \pmark & \cmark & \xmark & \xmark & \xmark & \dash & \cmark & \xmark \\
  \addlinespace
  \multicolumn{10}{@{}l}{\itshape Reconstruction, partition post hoc: structured-prior VAEs}\\
  GP-VAE \citep{Fortuin2020} & \xmark & \pmark & \cmark & \xmark & \xmark & \cmark & \xmark & \cmark & \xmark \\
  L-VAE \citep{Ramchandran2021} & \xmark & \cmark & \cmark & \cmark & \cmark & \cmark & \xmark & \cmark & \cmark \\
  LMM-VAE \citep{Ong2025} & \xmark & \cmark & \cmark & \cmark & \cmark & \cmark & \pmark & \cmark & \cmark \\
  \addlinespace
  \multicolumn{10}{@{}l}{\itshape Partition, reconstruction and class trajectories}\\
  DMLMM \citep{Kock2025} & \cmark & \cmark & \xmark & \xmark & \cmark & \cmark & \xmark & \cmark & \cmark \\
  LCMM \citep{ProustLima2017} & \cmark & \cmark & \xmark & \cmark & \cmark & \cmark & \xmark & \cmark & \cmark \\
  \textbf{LCMM-VAE (ours)} & \cmark & \cmark & \cmark & \cmark & \cmark & \cmark & \pmark & \cmark & \cmark \\
  \bottomrule
  \end{tabular*}\par\medskip
  {\raggedright\footnotesize
  \emph{Partition}: one class per subject; \pmark: one label per time step (SOM-VAE, T-DPSOM).
  \emph{Irregular times}: subject-specific observation times; \pmark: a regular grid with a missingness mask; \xmark: complete, equal-length series.
  \emph{High-dim.\ $Y$}: tens of markers in one fit; \pmark: univariate as published, concatenated here.
  \emph{Covariates}: trajectories depend on observed covariates.
  \emph{Random effects}: subject-specific deviations from the population or class trajectory.
  \emph{Generative}: a probabilistic model of the observations; \pmark: of the spline coefficients only.
  \emph{Identifiable}: \cmark: proven; \pmark: under assumptions the implementation does not meet; \xmark: no result; \dash: not applicable (no generative model).
  \emph{Imputation}: values at unobserved cells.
  \emph{Any $t$}: a subject or class trajectory at any time.\par}
  }
\end{minipage}
}]

\end{document}